\documentclass[sigconf]{acmart}
\AtBeginDocument{%
  }

\setcopyright{acmlicensed}
\copyrightyear{2018}
\acmYear{2018}
\acmDOI{XXXXXXX.XXXXXXX}
\acmConference[Conference acronym 'XX]{Make sure to enter the correct
  conference title from your rights confirmation email}{June 03--05,
  2018}{Woodstock, NY}
\acmISBN{978-1-4503-XXXX-X/2018/06}

\usepackage{booktabs}   
\usepackage{multirow}   
\usepackage{makecell}   
\usepackage[table]{xcolor}
\usepackage{arydshln}   

\usepackage{colortbl}
\definecolor{dqcolor}{RGB}{245,235,235}
\definecolor{edcolor}{RGB}{235,242,250}

\definecolor{uclablue}{RGB}{159, 195, 224}

\definecolor{uclagold}{RGB}{254,180,167}

\definecolor{grayred}{RGB}{232,237,205}

\newcommand{\nop}[1]{}
\definecolor{TealBlue}{rgb}{1.0, 0.97, 0.8}

\usepackage{pifont}
\newcommand{\cmark}{\textcolor{green!60!black}{\ding{51}}}
\newcommand{\xmark}{\textcolor{red}{\ding{55}}}

\usepackage{graphicx}
\usepackage[table]{xcolor}
\usepackage{geometry}

\definecolor{groupgray}{gray}{0.95}

\usepackage{tcolorbox}
\tcbuselibrary{skins,breakable}
\usepackage{enumitem}
\usepackage{twemojis}

\tcbset{
    dynamicbox/.style={
        enhanced,
        breakable,
        width=\linewidth,
        colback=white,
        colframe=gray!60!black,
        colbacktitle=gray!60!black,
        coltitle=white,
        fonttitle=\bfseries,
        title=#1,
        boxrule=0.8pt,
        left=4pt,
        right=4pt,
        top=4pt,
        bottom=4pt,
        toptitle=2pt,
        bottomtitle=2pt,
        lefttitle=4pt,
    }
}

\newcommand{\mytriangle}{$\blacktriangleright$}

\usepackage{amssymb} 
\usepackage{tcolorbox}
\usepackage{amssymb} 
\usepackage{tcolorbox}

\newtcolorbox{casebox}[2][]{
    colback=gray!5,      
    colframe=black!70,   
    title=\textbf{#2},   
    fonttitle=\bfseries\small,
    fontupper=\small,    
    boxrule=0.8pt,
    arc=2pt,
    width=\textwidth,    
    #1
}

\providecommand{\mytriangle}{$\blacktriangleright$}

\begin{document}
\begin{sloppy}

\title{Evaluating Long-Horizon Memory for Multi-Party Collaborative Dialogues}

\author{Chuanrui Hu}
\authornote{Equal contribution}
\email{chuanrui.hu@shanda.com}
\affiliation{%
  \institution{EverMind, Shanda Group}
  \country{USA}
}

\author{Tong Li}
\authornotemark[1]
\email{litong02@shanda.com}
\affiliation{%
  \institution{EverMind, Shanda Group}
  \country{USA}
}

\author{Xingze Gao}
\email{xingze.gao@shanda.com}
\affiliation{%
  \institution{EverMind, Shanda Group}
  \country{USA}
}

\author{Hongda Chen}
\email{hongda.chen@shanda.com}
\affiliation{%
  \institution{EverMind, Shanda Group}
  \country{USA}
}

\author{Yi Bai}
\email{baiyi@shanda.com}
\affiliation{%
  \institution{EverMind, Shanda Group}
  \country{USA}
}

\author{Dannong Xu}
\email{dannong.xu@shanda.com}
\affiliation{%
  \institution{EverMind, Shanda Group}
  \country{USA}
}

\author{Tianwei Lin}
\email{tianwei.lin@shanda.com}
\affiliation{%
  \institution{EverMind, Shanda Group}
  \country{USA}
}

\author{Xiaohong Li}
\email{xiaohong.li@shanda.com}
\affiliation{%
  \institution{EverMind, Shanda Group}
  \country{USA}
}

\author{Yunyun Han}
\email{hanyunyun@shanda.com}
\affiliation{%
  \institution{EverMind, Shanda Group}
  \country{USA}
}

\author{Jian Pei}
\email{j.pei@duke.edu}
\affiliation{%
  \institution{Duke University}
  \country{USA}
}

\author{Yafeng Deng}
\authornote{Contact author}
\email{dengyafeng@shanda.com}
\affiliation{%
  \institution{EverMind, Shanda Group}
  \country{USA}
}

\renewcommand{\shortauthors}{Hu et al.}
\begin{abstract}
Long-term conversational memory in practical LLM applications is inherently collaborative: information is produced by multiple participants, scattered across groups and channels, revised over time, and implicitly grounded in roles and social context. Yet there is currently no established benchmark that evaluates memory under interaction patterns resembling real-world deployment, as existing benchmarks largely focus on dyadic or single-topic dialogues. In this paper, we introduce \textbf{EverMemBench}, the first benchmark designed for long-horizon collaborative memory, built from multi-party, multi-group conversations spanning over one million tokens with dense cross-topic interleaving, temporally evolving decisions, and role-conditioned personas. EverMemBench evaluates memory systems using 2{,}400 QA pairs across three dimensions essential for real applications: \textit{fine-grained recall}, \textit{memory awareness}, and \textit{user profile understanding}. Our evaluation reveals fundamental limitations of current systems: multi-hop reasoning collapses under multi-party attribution even with oracle evidence (26\% accuracy), temporal reasoning fails without explicit version semantics beyond timestamps, and memory awareness is bottlenecked by retrieval, as similarity-based methods miss implicitly relevant information. EverMemBench thus represents a concrete step toward realistic evaluation of LLM memory and a cornerstone benchmark for developing next-generation LLMs that reason over time, roles, and collaborative interaction structure. Our benchmark and code are publicly available at \url{https://github.com/EverMind-AI/EverMemBench}.
\end{abstract}



\received{20 February 2007}
\received[revised]{12 March 2009}
\received[accepted]{5 June 2009}

\maketitle

\begin{table*}[t]
\centering
\footnotesize
\setlength{\tabcolsep}{4pt}
\renewcommand{\arraystretch}{1.1}

\begin{tabular}{l ccccc}
\toprule
\textbf{Aspects}
& \textbf{LoCoMo}
& \textbf{LongMemEval}
& \textbf{PersonaMem-v1}
& \textbf{PersonaMem-v2}
& \textbf{EverMemBench (Ours)} \\
\midrule

\multicolumn{6}{l}{\textit{\textbf{Dialogue Characteristics}}} \\
\midrule
Interaction Type & Dyadic & User--Assistant & User--Assistant & User--Assistant & \textbf{Multi-party Group} \\
Task Structure & Single-session & Long-term & Personalized & Preference-based & \textbf{Long-term Interdependent} \\
Dialogue Turns & 326.8 & 493.5 & 313.6 & 448.5 & \textbf{10,204.6} \\
Context Length & 9K & \textbf{1.5M} & 1M & 128K & 1M \\
Personas per Batch & 2 & 1 & 1 & 1 & \textbf{37.6} \\
\midrule

\multicolumn{6}{l}{\textit{\textbf{Dialogue Features}}} \\
\midrule
High-Info Dialog Flow & \xmark & \xmark & \xmark & \xmark & \cmark \\
Diverse Persona Interaction & \xmark & \xmark & \xmark & \xmark & \cmark \\
Cross-Topic Interaction & \xmark & \xmark & \xmark & \xmark & \cmark \\
User Knowledge Update & \xmark & \xmark & \cmark & \cmark & \cmark \\
\midrule

\multicolumn{6}{l}{\textit{\textbf{Evaluation Dimensions}}} \\
\midrule
Fine-Grained Recall & \cmark & \cmark & \cmark & \cmark & \cmark \\
Memory Awareness & \xmark & \xmark & \cmark & \cmark & \cmark \\
Profile Understanding$^\dagger$ & \xmark & \xmark & \xmark & \xmark & \cmark \\
\bottomrule
\end{tabular}

\vspace{2pt}
{\scriptsize $^\dagger$Profile Understanding denotes implicit user modeling from long-term dialogue, not explicit profile retrieval.}

\caption{Comparison with prior conversational memory benchmarks. EverMemBench uniquely supports multi-party group conversations with long-term interdependent tasks, high information density, and rich persona interaction.}
\vspace{-13pt}
\label{tab:memory_benchmarks}
\end{table*}

\section{Introduction}

Large language models are increasingly deployed as conversational agents in settings where interactions extend over time, span contexts, and involve multiple participants~\citep{brachman2025currentfutureuselarge,huang2025mempalmemorybasedpersonalizeddialogue,yi2025survey}. In practical applications such as workplace collaboration and personal assistance, conversational memory is inherently \emph{collaborative}: information is produced by different people, scattered across groups and channels, revised as decisions evolve, and implicitly shaped by roles and social relations. These settings impose two fundamental challenges for memory systems. First, conversations are often \textit{multi-party}, requiring the system to track who said what and how information propagates across speakers and groups~\citep{ganesh2023survey,gu2022says,sapkota2025multipartyconversationalagentssurvey}. Second, effective memory goes beyond verbatim recall, demanding the ability to retain fine-grained details for precise retrieval~\citep{maharana2024evaluatinglongtermconversationalmemory, wu2025longmemevalbenchmarkingchatassistants}, recognize when past information becomes relevant in new situations~\citep{wu2025interpersonal,samarinas2024procis}, and respond consistently with user preferences, expertise, and social context~\citep{zhaollms,zhao-etal-2025-personalens,jiang2025knowmerespondme}.

Despite rapid progress in long-context modeling and memory-augmented agents, evaluation has not kept pace. Many benchmarks implicitly equate stronger memory with the ability to process more tokens, treating memory as recall over long inputs~\citep{maharana2024evaluatinglongtermconversationalmemory,wu2025longmemevalbenchmarkingchatassistants,jiang2025knowmerespondme,tan2025membenchcomprehensiveevaluationmemory,nelson2024needlehaystackmemorybased}. In practice, however, failures rarely stem from context length alone~\citep{liu-etal-2024-lost,levy-etal-2024-task}. Instead, systems break down due to confusion about attribution in group chats, interference across closely related topics, inconsistency in persona and style, and inability to update beliefs when plans or constraints change. At the same time, recent memory-augmented systems~\citep{chhikara2025mem0buildingproductionreadyai,salama2025meminsightautonomousmemoryaugmentation,nan2025nemoriselforganizingagentmemory,li2025memosoperatingmemoryaugmentedgeneration,kang2025memoryosaiagent} increasingly attach persistent or structured memory to LLMs for long-horizon personalization and task continuity. Their growing deployment heightens the need for benchmarks that reflect realistic conversational dynamics, as it remains unclear which memory designs improve behavior under collaborative, evolving interactions rather than only boosting recall in constructed long contexts. This gap between existing benchmarks and practical memory demands is summarized in Table~\ref{tab:memory_benchmarks}.

A closer examination reveals that current benchmarks systematically underrepresent the structure of real interactions. Most focus on dyadic conversations~\citep{hu2025advancingmultipartydialogueframework}, whereas real-world settings involve multiple roles contributing to shared, interdependent decisions. Long contexts are often created by injecting topic-irrelevant distractors~\citep{kamradt2023needle,chen2024documenthaystacksvisionlanguagereasoning,nelson2024needlehaystackmemorybased}, which tests noise tolerance but not relevance recognition in coherent, interleaved dialogues. Persona modeling is typically shallow, failing to capture how communication style and expertise emerge from role relations and repeated interaction~\citep{zhang-etal-2018-personalizing,jiang2025knowmerespondme}. Finally, many benchmarks assume stationary facts, while real conversational memory must support explicit updates, revisions, and conflict resolution as information evolves over time.

\begin{figure*}[t]
  \centering
  \includegraphics[width=0.8\textwidth]{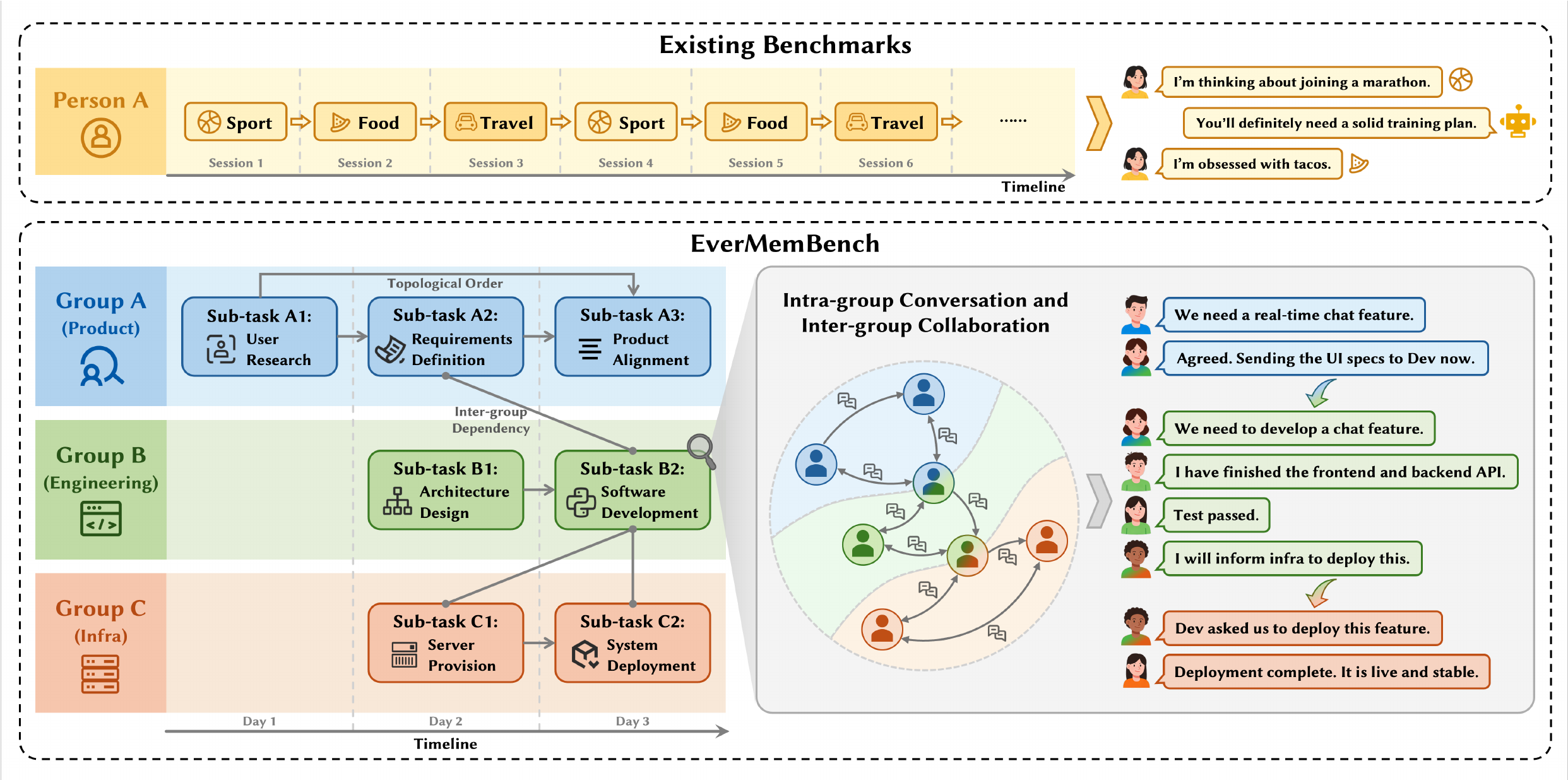}
  \caption{Existing benchmarks vs.\ EverMemBench. Existing benchmarks focus on dyadic, single-topic sessions. EverMemBench models multi-party collaboration across interdependent groups, where information is distributed across speakers, channels, and time, requiring cross-group reasoning and temporal tracking absent in dyadic settings.}
  \Description{teaser}
  \label{teaser}
\end{figure*}

To bridge this gap, we introduce \textbf{EverMemBench}, a benchmark designed to evaluate long-horizon memory under interaction patterns that closely resemble real-world deployment. EverMemBench is built from information-dense dialogues in which multiple roles participate across interconnected group chats, exhibiting coherent cross-topic interleaving and revisiting of earlier decisions rather than isolated topic sessions. It features diverse personas with role-conditioned skills and communication styles, and models dynamic user knowledge where earlier information can be revised as constraints change. A high-level comparison with prior benchmarks is shown in Figure~\ref{teaser}.

To systematically assess these challenges, EverMemBench evaluates memory systems along three dimensions that are essential for collaborative assistants: \textbf{fine-grained recall} for accurately retrieving specific entities from dense, multi-party discussions, \textbf{memory awareness}  for comprehending stored knowledge and applying it to novel, unseen scenarios, and \textbf{profile understanding} for maintaining consistency with user preferences, expertise, and roles. Experiments on both long-context LLMs and memory-augmented systems reveal persistent limitations across all three dimensions. 

In summary, this paper makes three contributions. First, we introduce \textbf{EverMemBench}, the first benchmark explicitly designed to evaluate long-horizon memory in multi-party, multi-group conversational settings, featuring information-dense 
dialogues across five projects, each spanning one million tokens, with coherent cross-topic interaction, role-conditioned personas, and dynamic knowledge updates that mirror real-world collaboration. Second, we propose three evaluation dimensions—\textbf{fine-grained recall}, \textbf{memory awareness}, and \textbf{profile understanding}—that directly capture the core capabilities required for effective collaborative memory, moving beyond token-level recall to relevance recognition and persona consistency. Third, through systematic experiments on both long-context LLMs and memory-augmented systems, we reveal persistent structural limitations: multi-hop reasoning collapses under multi-party attribution, temporal reasoning degrades as topic interleaving obscures event lifecycle boundaries (initiation, completion, archival),
and similarity-based memory retrieval fails to surface implicitly relevant information. Together, these contributions position EverMemBench as a concrete step toward realistic evaluation of LLM memory and a cornerstone benchmark for developing next-generation LLMs that reason over time, roles, and collaborative interaction structure.


\begin{table*}[t]
\footnotesize
\setlength{\tabcolsep}{5pt}
\renewcommand{\arraystretch}{1.15}
\begin{tabular}{@{} >{\raggedright\arraybackslash}p{1.8cm} p{\dimexpr\textwidth-1.8cm-4\tabcolsep\relax} @{}}
\toprule
\multicolumn{2}{@{}l}{\textbf{Fine-grained Recall}} \\
\midrule
Single-hop 
  & Retrieves precise entities while filtering out semantically similar distractors (e.g., intermediate drafts). \\
 Retrieval & \textit{\textcolor{blue}{\textbf{Q:}} What’s the link to Person Q’s final deliverable for Task A?}
    \quad \textcolor{green}{\textcolor{green}{$\checkmark$}} Confluence link
    \quad \textcolor{red}{$\times$} Figma link (same speaker, same task, 2 days earlier) \\[3pt]
Multi-hop 
  &  Traces an individual’s work timeline across fragmented threads and groups via multi-hop reasoning \\
Trajectory  & \textit{\textcolor{blue}{\textbf{Q:}} What is the next task assigned to the owner of Task~A after completing Task~A?}
    \quad \textcolor{green}{\textbf{$\checkmark$}} Task~B
    \quad \textcolor{red}{$\times$} Task~ C\,(Finds the person but fails to infer the next step)\\[3pt]
Temporal 
  & Extracts time spans from noisy contexts with identical phrases and adjacent dates. \\
  Duration & \textit{\textcolor{blue}{\textbf{Q:}} How many days from Task~A start to archival?}
    \quad \textcolor{green}{\textbf{$\checkmark$}} 7 days
    \quad \textcolor{red}{$\times$}\, 264 days (find the wrong anchor) \\
\midrule
\multicolumn{2}{@{}l}{\textbf{Memory Awareness}} \\
\midrule
Constraint
  &  Apply implicit organizational norms and constraints from prior memories to guide decisions in an \textbf{generalize} scenario\\
  & \textit{\textcolor{blue}{\textbf{Q:}} A new field must be added to the shared schema—who should own the change?}
    \quad \textcolor{green}{\textbf{$\checkmark$}} Dev A (10+ msgs related, definer)
    \quad \textcolor{red}{$\times$} Dev~B (30+ msgs, consumer) \\[3pt]
Proactivity
  & Proactively recall explicit rules and detect conflicts when an unseen task instruction would violate them, remaining robust to leading or persuasive framing.\\
  &   \textit{\textcolor{blue}{\textbf{Q:}} The customer will sign today if we offer 25\% off—draft the contract now!!}
    \quad \textcolor{green}{\textbf{$\checkmark$}}  Remaider: 25\% off needs pre-approval; 
    \quad \textcolor{red}{$\times$} Draft  the contract \\[3pt]
Update
  & Tracks rule evolution, composing base protocols with later overrides to apply the updated policy. \\
   & \textit{\textcolor{blue}{\textbf{Q:}} For a \emph{new} service created after the 2026-01-15 policy update, which CI should we use?} 
    \quad \textcolor{green}{\textbf{$\checkmark$}} GitHub Actions (current policy) 
    \quad \textcolor{red}{$\times$} Jenkins (old handbook) \\ 
\midrule
\multicolumn{2}{@{}l}{\textbf{Profile Understanding}} \\
\midrule
Style
  & Generate a personalized response that matches an individual’s implicit communication style inferred from prior dialogues (tone/structure/verbosity).\\
  & \textit{\textcolor{blue}{\textbf{Q:}} Draft a project update on this user's behalf.}
    \quad \textcolor{green}{\textbf{$\checkmark$}} Terse bullets with jargon (matches user)
    \quad \textcolor{red}{$\times$} Formal paragraphs (generic  tone) \\[3pt]
Skill
  & Infer and apply a persona’s competence boundary from long-term memory, producing recommendations that match what the speaker would realistically propose (and rejecting overly generic best-practice suggestions). \\
  & \textit{\textcolor{blue}{\textbf{Q:}} How should person A  optimize this service quickly?}
   \quad \textcolor{green}{\textbf{$\checkmark$}} JVM profiling + GC tuning (Java)
  \quad \textcolor{red}{$\times$}  Use pandas tooling (Pythons, beyond A’s capabilitie)\\[3pt]
Role
  & Adopt the speaker’s professional role perspective when responding. \\
  & \textit{\textcolor{blue}{\textbf{Q:}} Write a post-mortem for the service outage.}
    \quad \textcolor{green}{\textbf{$\checkmark$}} User impact, SLA breach, process gaps (PM)
    \quad \textcolor{red}{$\times$} Memory-leak fix, GC tuning (engineer) \\
\bottomrule
\end{tabular}
\captionsetup{skip=1.5pt}
\caption{Overview of the nine tasks with illustrative mini-cases.
Full cases with evidence chains are in Appendix~\ref{appendix:examples}.}
\label{tab:task_overview}
\end{table*}

\section{Related Work}

\paragraph{Long-Context Conversational Memory Benchmarks.}
Recent benchmarks evaluate long-horizon conversational memory from multiple perspectives, including long multi-session interaction, capability-factorized assistant memory, and dynamic personalization. LoCoMo~\citep{maharana2024evaluatinglongtermconversationalmemory} focuses on very long multi-session conversations with tasks such as question answering and event summarization. LongMemEval~\citep{wu2025longmemevalbenchmarkingchatassistants} decomposes chat-assistant memory into abilities including information extraction, multi-session and temporal reasoning, knowledge updates, and abstention. PersonaMem~\citep{jiang2025knowmerespondme} emphasizes dynamic user profiling and personalization over extended user--LLM histories. More recent efforts further broaden coverage of memory abilities and dialogue settings, including MemBench~\citep{tan2025membenchcomprehensiveevaluationmemory}, MADial-Bench~\citep{he2024madialbenchrealworldevaluationmemoryaugmented}, and BEAM~\citep{tavakoli2025milliontokensbenchmarkingenhancing}.

Despite their differences, these benchmarks share a simplifying assumption: interactions are dyadic or centered on a single user and a single assistant. Even when multiple speakers exist, interdependence across roles and groups and explicit attribution reasoning are typically not treated as first-class evaluation targets. This assumption fundamentally simplifies attribution, relevance, and update reasoning, and consequently under-stresses collaborative phenomena that dominate real applications, such as dense multi-party attribution, coherent cross-topic interleaving, persona shifts under social context, and non-stationary user knowledge that must be revised and reconciled. EverMemBench departs from this paradigm by explicitly targeting multi-party group chat with high information density, interdependent tasks, diverse personas shaped by role relations, and evolving knowledge states, together with tasks that disentangle detailed recall, memory awareness, and user profile understanding.

\paragraph{Memory-Augmented Systems and Architectures.}
Memory-augmented systems increasingly treat memory as an explicit component that can be persisted, structured, retrieved, and updated, with designs ranging from pragmatic memory layers to more autonomous organization and OS-level abstractions. Mem0~\citep{chhikara2025mem0buildingproductionreadyai} and MemInsight~\citep{salama2025meminsightautonomousmemoryaugmentation} propose scalable persistent memory layers that extract salient information from conversational histories and retrieve it when needed. Zep~\citep{rasmussen2025zeptemporalknowledgegraph} takes a different approach, building memory on a temporal knowledge graph whose bi-temporal model tracks both event time and ingestion time, enabling conflict resolution as facts evolve. MemoBase~\citep{memobase2025} organizes memory as structured user profiles with developer-defined schemas, prioritizing user-centric personalization over general-purpose retrieval. Beyond persistent stores, A-MEM~\citep{xu2025amemagenticmemoryllm} frames memory as an agentic module that decides what to store and how to use it, while Nemori~\citep{nan2025nemoriselforganizingagentmemory} proposes self-organizing memory emphasizing structured organization and evolution. Another line elevates memory to an infrastructure abstraction: MemOS~\citep{li2025memosoperatingmemoryaugmentedgeneration} and MemoryOS~\citep{kang2025memoryosaiagent} treat memory as an OS-like resource, aiming to unify heterogeneous memories with scheduling and lifecycle management. Across these designs, retrieval-augmented generation commonly serves as the backbone for accessing external or long-range information~\citep{chen2024documenthaystacksvisionlanguagereasoning,borgeaud2022improvinglanguagemodelsretrieving,yang2025wikiautogenmultimodalwikipediastylearticle,lewis2021,guu2020realmretrievalaugmentedlanguagemodel,yang2025inexhallucinationmitigationintrospection}.

While architecturally diverse, these systems are predominantly evaluated in dyadic or single-user settings that obscure the challenges they are designed to address. Such evaluations under-stress multi-party attribution, cross-group dependency, temporal revision, and persona consistency under shifting social context, making it difficult to distinguish memory mechanisms that truly support realistic collaboration from those that succeed only under simplified conditions. EverMemBench provides a complementary and more diagnostic evaluation environment, stressing memory architectures under collaborative interaction where information is distributed across speakers, groups, and time, and enabling systematic analysis of how different designs succeed or fail in practice.

\section{EverMemBench}

\nop{
\begin{figure*}[t!]
	\begin{center}
        \includegraphics[width=\textwidth]{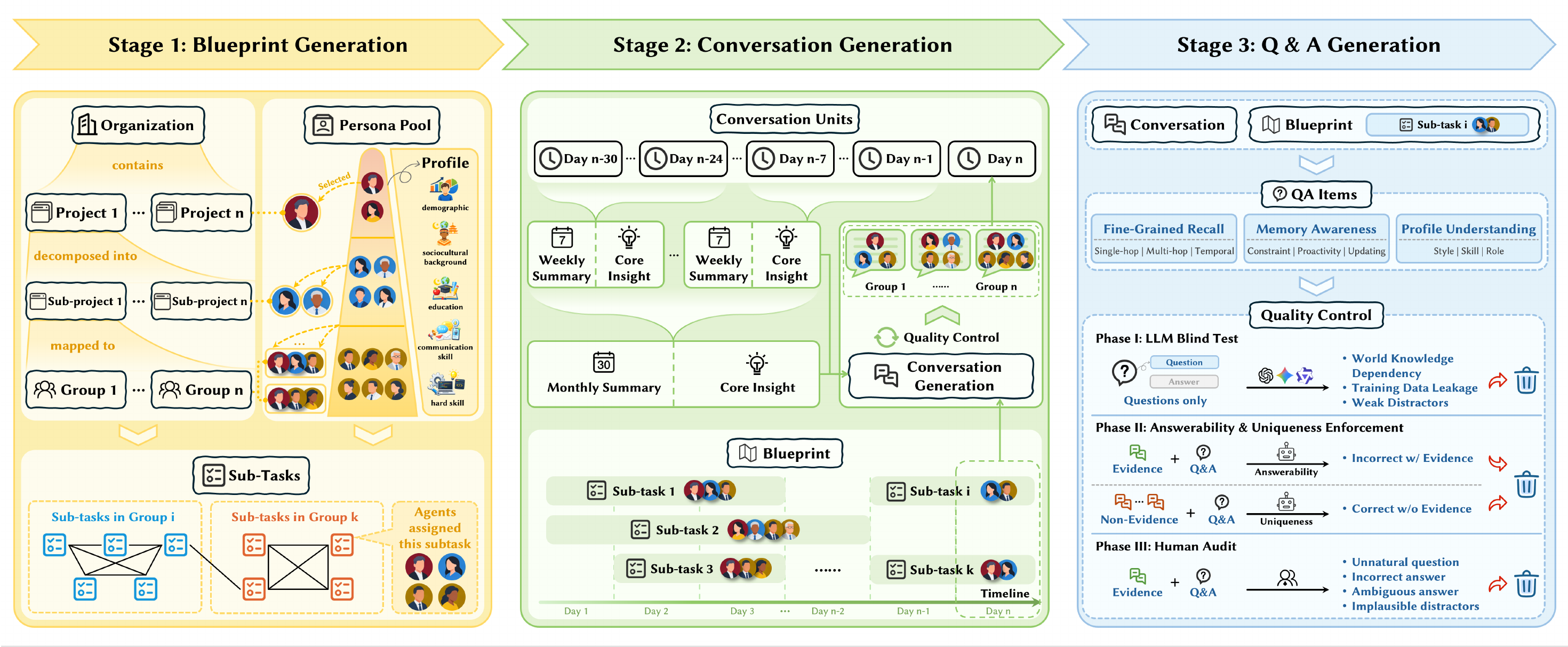}
	\end{center}
    \vspace{-10pt}
    	\caption{\textbf{Data curation pipeline of EverMemBench.}
Stage~1 builds organizational structure, persona profiles, and sub-task assignments;
Stage~2 generates daily dialogues conditioned on hierarchical summaries;
Stage~3 produces QA pairs with three-phase quality control.}
	\label{curation}
     \vspace{-10pt}
\end{figure*}

\subsection{Evaluation Dimensions}

EverMemBench aims to evaluate LLMs as \textbf{long-term collaborators} in complex, dynamic environments. Unlike prior dyadic benchmarks~\citep{maharana2024evaluatinglongtermconversationalmemory,wu2025longmemevalbenchmarkingchatassistants,jiang2025knowmerespondme,jiang2025personamemv2personalizedintelligencelearning}, real-world collaboration requires navigating dense multi-party coordination, adapting to evolving constraints, and maintaining role-dependent consistency. We instantiate these challenges in a high-fidelity workplace simulation, as it naturally concentrates the core difficulties of realistic memory.

To systematically diagnose these capabilities, we decompose memory competence into three complementary task families:

\begin{itemize}
    \item \textbf{Fine-grained Recall} evaluates direct memory lookups, anchoring queries to specific factual details with definitive answers derived from stored records. Sub-tasks include: \textit{Single-hop} for direct entity retrieval; \textit{Multi-hop} for cross-group reasoning across personas or channels; and \textit{Temporal} for chronological reasoning about version histories.

    \item \textbf{Memory Awareness} tests whether the system can mobilize existing memories to solve problems in unseen situations, differentiating active comprehension from passive storage. Sub-tasks include: \textit{Constraint} for applying implicit rules from prior conversations; \textit{Proactivity} for surfacing relevant context when evaluating proposed actions; and \textit{Updating} for identifying current valid states when earlier decisions have been superseded.

    \item \textbf{Profile Understanding} assesses the ability to maintain distinct user personas by mining implicit habits and long-term traits from dialogue history. Sub-tasks include: \textit{Style} for matching communication patterns; \textit{Skill} for determining responses based on professional expertise; and \textit{Role} for inferring focus areas from organizational role.
\end{itemize}

\subsection{Task Formulation}
We evaluate memory systems via a streaming multi-group protocol. 
The benchmark comprises 5 \textit{projects} spanning diverse domains to assess cross-domain adaptability; each project maintains isolated dialogue history, memory state, and evaluation queries. Within each project, $N$ groups engage in a simulated year of daily conversations. During the \textbf{ingestion phase}, the system receives \textit{raw multi-party conversations} organized by day and group: on each day $d$, it observes the daily batch $\{M_{d,g}\}_{g=1}^{N}$, where $M_{d,g}$ denotes the chronologically ordered multi-party message list of group $g$ on day $d$. The system must autonomously process these interleaved multi-party exchanges to build and maintain its memory state. After full ingestion, the \textbf{evaluation phase} poses queries against the complete history. Queries take two forms: (1) \textit{multiple-choice} questions scored by exact option matching, and (2) \textit{open-ended} questions scored by an LLM judge. Each question targets one of the three evaluation dimensions and is annotated with evidence spans for diagnostic analysis.

\subsection{Data Construction}
Generating coherent long-horizon multi-party dialogues poses three key challenges: 
(1) context-window limits force truncation of early history; 
(2) logical inconsistencies emerge as conversations grow; 
(3) persona and temporal coherence degrade over extended interactions. 
To address these, we decompose data construction into three stages (Figure~\ref{curation}): 
\textbf{Stage~1} generates a global \textit{blueprint} encoding organizational structure, 
individual persona profiles, and a project timeline with main and auxiliary storylines; 
\textbf{Stage~2} synthesizes dialogues in chunks conditioned on the blueprint; 
\textbf{Stage~3} generates evidence-grounded QA pairs from the dialogues and blueprint. 
This separation ensures global consistency while enabling scalable, chunk-wise generation.
All generation processes (blueprint, dialogue, and QA) employ Gemini-2.5-Pro~\citep{comanici2025gemini25pushingfrontier} as the backbone model.
\subsubsection{\textbf{Preliminaries}}
We first establish the shared foundation for data generation.
We predefine an organizational skeleton $S$ comprising 170 employees $\mathcal{E}=\{e_1,\dots,e_{170}\}$ across 7 departments, structured in a three-tier hierarchy (1 CEO, 7 directors, 164 staff), and five projects about different domains $\mathcal{P}=\{p_1,\dots,p_5\}$.
Given $S$, we utilize LLM to generate a persona profile for each employee:
\begin{equation}
\pi_e = \big(\,\mathrm{rank}_e,\; \mathrm{dept}_e,\; \mathrm{role}_e,\; \mathbf{s}_e,\; \mathbf{c}_e\,\big),
\end{equation}
where $\mathbf{s}_e$ is a skill set (40--60 competencies) and $\mathbf{c}_e$ is an 8-dimensional communication style vector.


\subsubsection{\textbf{Blueprint Generation}}

For each project $p\in\mathcal{P}$, we construct an independent blueprint $B_p$ that guides all subsequent dialogue generation.
The blueprint encodes:
\begin{equation}
B_p = \big(\,\mathcal{E}_p,\; \{(\mathcal{E}_{p,j},\, \mathcal{T}_{p,j})\}_{j=1}^{3}\,\big),
\end{equation}
where $\mathcal{E}_p\subset\mathcal{E}$ is the project team (a subset of 20--60 employees selected from the full pool), and for each of the three sub-projects $j\in\{1,2,3\}$: $\mathcal{E}_{p,j}\subseteq\mathcal{E}_p$ is the assigned member set (with possible overlaps across sub-projects), and $\mathcal{T}_{p,j}$ is the set of sub-tasks generated based on member profiles and sub-project goals.
Together with the shared foundation $(S, \Pi)$, the blueprint $B_p$ enforces decision-logic consistency, persona consistency, and temporal coherence during dialogue generation.

We construct $B_p$ through the following steps:
(1) select a project team $\mathcal{E}_p$ (20--60 members) from the employee pool via skill matching;
(2) decompose project $p$ into 3 concurrent sub-projects;
(3) assign members to each sub-project $\mathcal{E}_{p,j}$, where one member may participate in 1--3 sub-projects;
(4) generate sub-tasks $\mathcal{T}_{p,j}$ for each sub-project based on member profiles and project requirements.

\subsubsection{\textbf{Conversation Generation}}
Dialogues are generated incrementally for each project $p$ over a simulated timeline of $D$ days.
Each sub-project $j\in\{1,2,3\}$ maintains a dedicated group chat; however, since all three serve the same parent project, cross-group information sharing naturally emerges through overlapping membership and inter-dependent sub-tasks.

A key challenge is maintaining coherence over hundreds of days while respecting LLM context limits.
We address this through hierarchical summarization.
Let $C_{p,j}^{(<d)}=\{C_{p,j}^{(1)},\dots,C_{p,j}^{(d-1)}\}$ denote the conversation history of sub-project $j$ before day $d$.
We compute:
\begin{align}
W_{p,j}^{(d)} &= \mathrm{Summarize}\big(C_{p,j}^{(d-7:d-1)},\; \mathcal{T}_{p,j}^{(d-7:d-1)}\big), \\
M_{p,j}^{(d)} &= \mathrm{Summarize}\big(C_{p,j}^{(d-30:d-1)},\; \mathcal{T}_{p,j}^{(d-30:d-1)}\big), \\
L_{p,j}^{(d)} &= \mathrm{Extract}_{\text{leader}}\big(C_{p,j}^{(d-1)}\big),
\end{align}
where $W_{p,j}^{(d)}$ is a weekly summary capturing recent progress, $M_{p,j}^{(d)}$ is a monthly summary preserving high-level milestones, and $L_{p,j}^{(d)}$ contains leader instructions (e.g., task assignments) extracted from the previous day.
This multi-scale context compresses unbounded history into a fixed-size window without losing temporal structure.

To avoid error accumulation from turn-by-turn decoding and to enable the model to plan coherent multi-party dynamics holistically, we generate each group's full daily conversation in a single pass.
Given the scheduled sub-tasks $\mathcal{T}_{p,j}^{(d)}$ for day $d$, the dialogue is synthesized as:
\begin{equation}
C_{p,j}^{(d)} = \mathrm{LLM}_{\text{dialog}}\Big(\mathcal{T}_{p,j}^{(d)},\; W_{p,j}^{(d)},\; M_{p,j}^{(d)},\; L_{p,j}^{(d)},\; \{\pi_e\}_{e\in\mathcal{E}_{p,j}}\Big).
\end{equation}

Since LLMs inevitably exhibit hallucination~\citep{Huang_2025}, we apply a three-stage quality verification to each generated dialogue to ensure data reliability.
First, a \textit{logic check} verifies that the dialogue is consistent with the sub-task specification---correct participants are involved and task details are accurately discussed.
Second, a \textit{profile check} ensures that each utterance aligns with the speaker's persona, including communication style and domain expertise.
Third, a \textit{progress check} validates that task completion status matches the scheduled timeline---tasks should neither finish prematurely nor be delayed without justification.
Dialogues failing any check are regenerated until all constraints are satisfied.
After acceptance, leader instructions $L_{p,j}^{(d+1)}$ are extracted for the next day, and the process repeats until the project concludes.
The complete conversation corpus for project $p$ is then:
\begin{equation}
\mathcal{C}_p = \big\{C_{p,j}^{(d)} : j\in\{1,2,3\},\; d\in\{1,\dots,D\}\big\}.
\end{equation}

\subsubsection{\textbf{Q\&A Generation}}\label{sec:qa_gen}

We construct evaluation items as QAE triples $(q, a, e)$, where $q$ is a question, $a$ is the answer, and $e \subset \mathcal{C}_p$ denotes the evidence spans from the dialogue corpus that support the answer. We generate QA pairs through three specialized pipelines corresponding to the three evaluation dimensions, each employing task-specific strategies for both question generation and distractor design.

\paragraph{Fine-grained Recall}
We adopt an \textbf{outline-driven protocol} combining \textit{structure mining} and \textit{contextual injection}. Given blueprint $B_p$ and dialogue corpus $\mathcal{C}_p$, we traverse the task structure to identify reasoning patterns (e.g., cross-group information chains, temporal update sequences) and extract associated dialogue spans as evidence $e$. An LLM then generates $(q, a)$ pairs grounded in $e$. For scenarios requiring constraints absent in the base narrative, we apply \textit{non-conflicting implantation}: supplementary evidence derived from $B_p$ is embedded into $\mathcal{C}_p$ without altering existing content, enabling controlled probing of corner cases while preserving conversational coherence.

\paragraph{Memory Awareness}
We adopt a \textbf{scenario-construction pipeline}. For \textit{Updating}, we locate state transitions in $\mathcal{C}_p$ where new information supersedes previous states; distractors represent stale information. For \textit{Constraint}, we extract implicit rules (e.g., ``always use Python for data tasks'') and construct scenarios requiring their application; distractors involve incorrect rules or constraint violations. To increase difficulty, we apply \textbf{adversarial perturbation}: \textit{keyword substitution} removes surface-level cues, and \textit{parameter removal} strips explicit triggers, forcing reliance on semantic understanding over lexical matching.

\paragraph{Profile Understanding}
We extract user traits from interaction patterns in $\mathcal{C}_p$ and generate questions requiring inference of these traits. Distractors are designed to isolate genuine comprehension from heuristic shortcuts. For \textit{Style}, we construct an \textbf{adversarial matrix} with options varying along two dimensions (Fact Correctness $\times$ Style Match); this $2\times 2$ design separates style comprehension from factual accuracy, with deliberate length-bias in incorrect options. For \textit{Skill} and \textit{Role}, distractors are generated via reasoning chains to be ``plausible but wrong'' (e.g., attributing design tool usage to a backend engineer), testing understanding of role boundaries and professional expertise.

\subsubsection{\textbf{Q\&A Quality Control}}\label{sec:quality_control}

We implement a three-phase filtering pipeline to ensure benchmark validity: samples must require memory access and be uniquely grounded in specific evidence.

\paragraph{Phase I: Blind Test}
A sample $(q, a)$ is discarded if an LLM can predict $a$ given only $q$ without any dialogue context. This filters out: (1) \textit{parametric memorization}, i.e., answers derivable from world knowledge in model weights; and (2) \textit{annotation artifacts}, i.e., low-quality distractors that make correct answers guessable.

\paragraph{Phase II: Evidence Grounding}
We partition $\mathcal{C}_p$ into segments $\mathbf{S} = \{S_1, \dots, S_n\}$ based on dialogue subtasks. For each $(q, a, e)$, let $S^+$ denote the segment containing evidence $e$, and $S^- = \mathbf{S} \setminus \{S^+\}$. We enforce dual verification:
\begin{itemize}
    \item \textbf{Sufficiency}: LLM must correctly derive $a$ from $(q, S^+)$, ensuring $e$ is self-contained.
    \item \textbf{Uniqueness}: LLM must fail to derive $a$ from $(q, S_j)$ for all $S_j \in S^-$, ensuring no unintended solution paths.
\end{itemize}

\paragraph{Phase III: Human Audit}
Expert annotators review samples surviving automated filters for: (1) \textit{logical incoherence}, i.e., structurally valid but semantically nonsensical questions; and (2) \textit{pragmatic implausibility}, i.e., distractors trivially dismissible via common sense despite satisfying formal constraints.

\subsection{Data Statistics}
\begin{figure}[t]
\centering
\includegraphics[width=.8\linewidth]{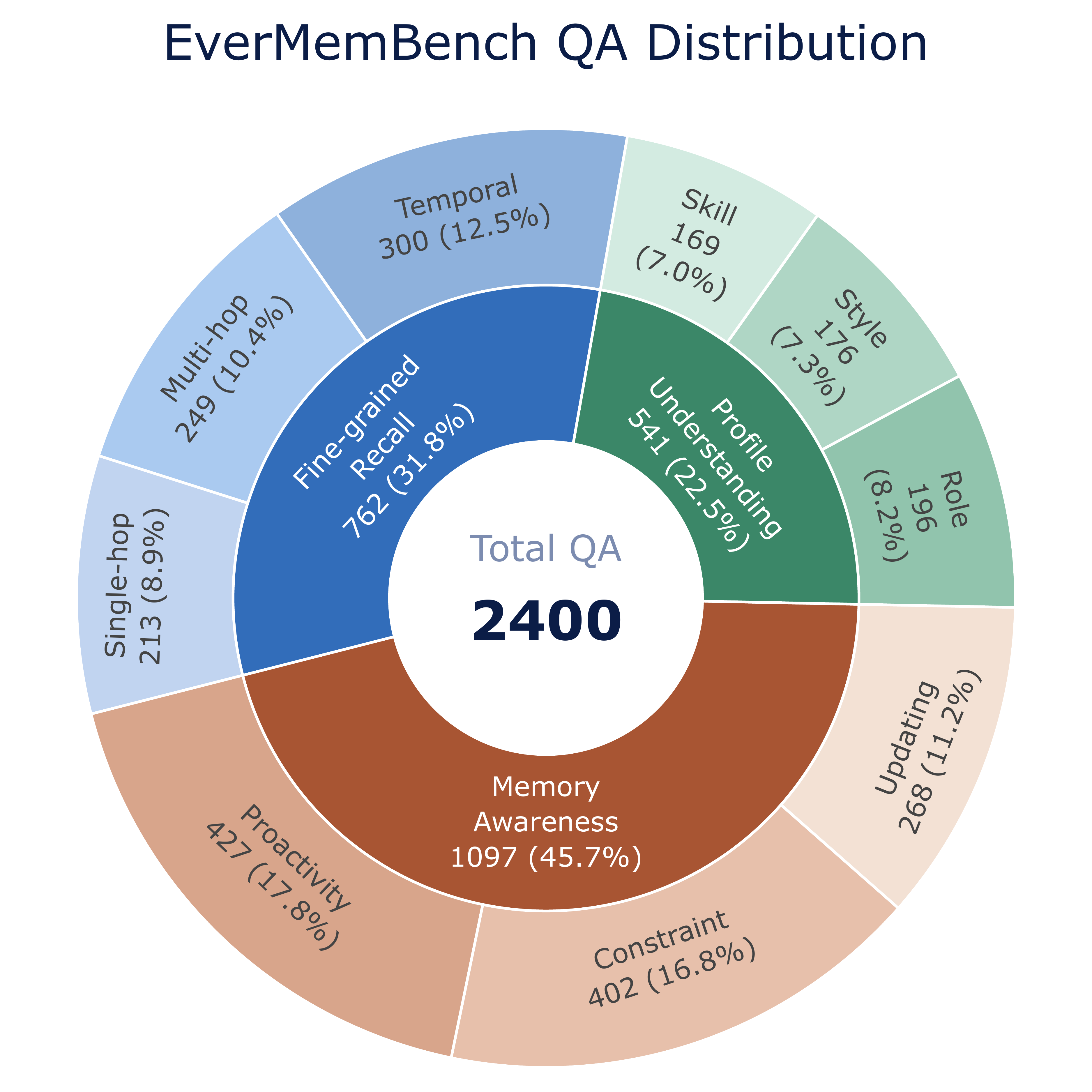}
\caption{Distribution of QA pairs.}
\label{fig:qa_distribution}
\end{figure}
\begin{table}[t]
\centering
\small
\setlength{\tabcolsep}{3.5pt}
\begin{tabular}{l r}
\toprule
\textbf{Statistic} & \textbf{Value} \\
\midrule
\multicolumn{2}{l}{\textit{Organizational Structure}} \\
\quad \# Projects / Sub-projects & 5 / 15 \\
\quad \# Employees & 170 \\
\quad \quad Executive / Manager / Staff & 1 / 5 / 164 \\
\quad Avg \# Participants / Project & 37.6 \\
\midrule
\multicolumn{2}{l}{\textit{Dialogue Statistics}} \\
\quad Time Span (days) & 365 \\
\quad Total Dialogue Turns & 51,023 \\
\quad Total Tokens & 4,225,555 \\
\quad Avg Tokens / Project & 845,111 \\
\quad Avg Turns / Day & 28.0 \\
\quad Avg Tokens / Turn & 82.8 \\
\midrule
\multicolumn{2}{l}{\textit{Evaluation Statistics}} \\
\quad \# QA Pairs & 2,400 \\
\quad \# Evaluation Dimensions & 3 \\
\quad \# Sub-dimensions & 9 \\
\bottomrule
\end{tabular}
\caption{\textbf{Data statistics of EverMemBench.}}
\label{tab:data_stats}
\vspace{-13pt}
\end{table}

EverMemBench targets multi-party collaborative memory in realistic workplace settings, a substantially more challenging scenario than dyadic or single-user benchmarks.
As summarized in Table~\ref{tab:data_stats}, the benchmark comprises 5 projects spanning diverse domains (Technology, Operations, Marketing, Financial Services, and Governance), each containing 3 interdependent sub-project groups with a simulated timeline of 365 days.
In total, 170 employees participate across projects, producing 51,023 dialogue turns and 4.2M tokens, with each project averaging approximately 1M tokens.

A key distinction from prior benchmarks is information density.
Existing long-context benchmarks often achieve large context windows by padding with topic-irrelevant distractors~\citep{wu2025longmemevalbenchmarkingchatassistants}. 
In contrast, EverMemBench generates complete, coherent event sequences through blueprint-guided synthesis, packing over 10,000 turns into each 1M-token project with the majority of events directly utilized for question design.

Figure~\ref{fig:qa_distribution} shows the distribution of 2,400 QA pairs across 3 dimensions and 9 sub-tasks, with balanced coverage over the five project topics (467--493 per topic).

}

\begin{figure*}[t!]
	\begin{center}
        \includegraphics[width=\textwidth]{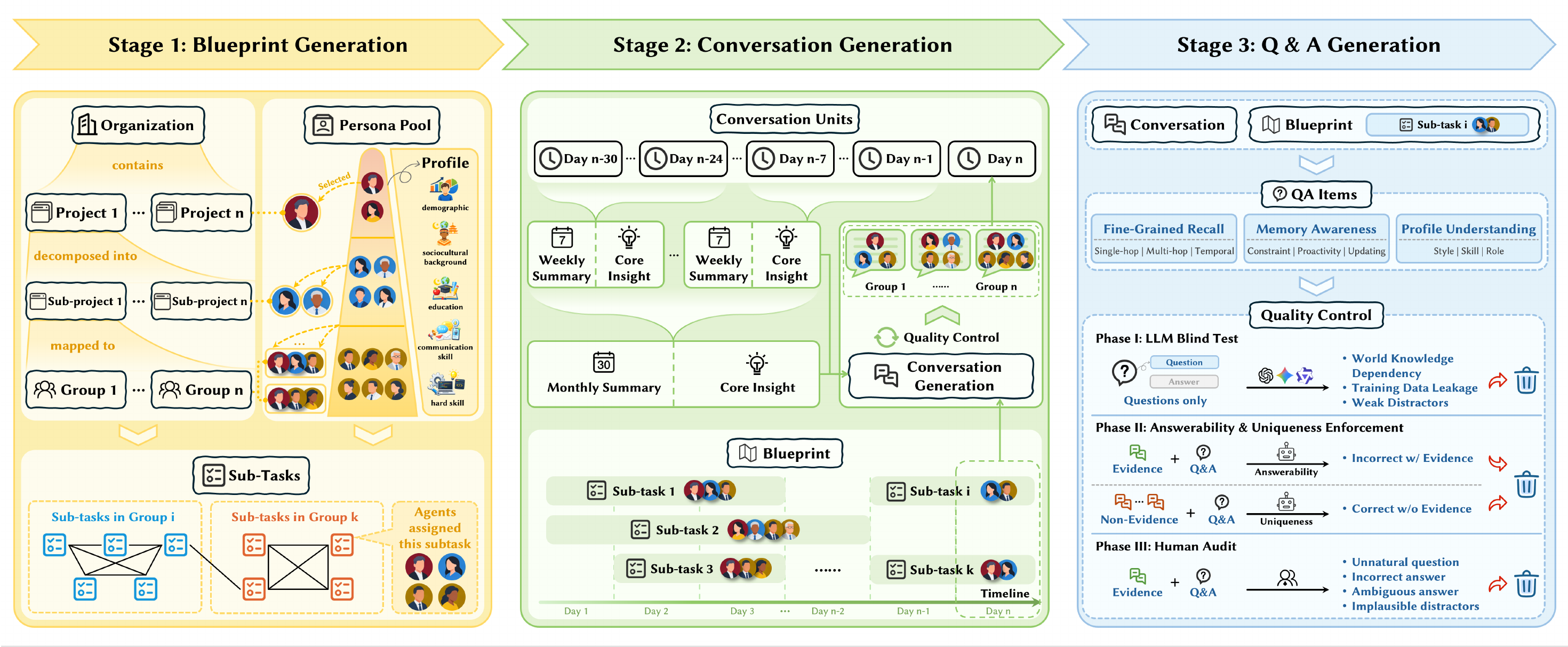}
	\end{center}
    \vspace{-10pt}
    	\caption{\textbf{Data curation pipeline of EverMemBench.}
Stage~1 builds organizational structure, persona profiles, and sub-task assignments;
Stage~2 generates daily dialogues conditioned on hierarchical summaries;
Stage~3 produces QA pairs with three-phase quality control.}
	\label{curation}
     \vspace{-10pt}
\end{figure*}

EverMemBench is designed not merely as a synthetic data artifact, but as a \emph{diagnostic instrument}: each design choice is aimed at exposing the structural failure modes that plague deployed conversational agents. Concretely, we build multi-party, multi-group dialogues with explicit temporal structure, role-conditioned personas, and tightly coupled tasks so that failures in attribution, temporal revision, and inferential retrieval become visible and measurable. Below we summarize the benchmark's evaluation goals, the streaming task protocol, and the three-stage curation pipeline that guarantees coherence, controllability, and evidentiary grounding.

\subsection{Evaluation Dimensions}

We evaluate LLMs as \textbf{long-term collaborators} in settings where failures are not caused by token limits alone but by interaction structure. Prior dyadic benchmarks miss these failure modes because they hide attribution and revision complexity behind single-threaded dialogs~\citep{maharana2024evaluatinglongtermconversationalmemory,wu2025longmemevalbenchmarkingchatassistants,jiang2025knowmerespondme,jiang2025personamemv2personalizedintelligencelearning}. To make the missing challenges explicit, EverMemBench defines three complementary task families that together capture the core competencies required for realistic collaboration.

\textbf{Fine-grained Recall} measures whether a system can retrieve precise facts from dense, multi-turn discussions where relevant evidence is scattered across speakers and groups. We include \textit{Single-hop} (local grounding and entity disambiguation), \textit{Multi-hop} (cross-group chaining), and \textit{Temporal} (event boundary point identification) 
sub-tasks specifically because these are the concrete operations systems must perform in practice to, e.g., determine the \emph{final} approved budget or the \emph{actual} assignee after several revisions.

\textbf{Memory Awareness} tests whether a system can move beyond simple retrieval to reason over stored information and apply it to novel scenarios. Our benchmark focuses on three memory-aware behaviors: Constraint (norm generalization), Proactivity (conflict detection under potentially biased instructions), and Update (rule versioning and precedence). Together, these sub-tasks assess whether an assistant can comprehend, revise, and proactively apply its memories in situations it has never encountered—capabilities that are essential for deployed conversational assistants.


\textbf{Profile Understanding} examines whether a system aggregates distributed signals into stable user models that guide behavior. We evaluate \textit{Style} (communication patterns), \textit{Skill} (expertise-based choices), and \textit{Role} (role focus) because real assistants must adapt tone, suggested actions, and assumptions based on inferred roles--capabilities that cannot be verified by single-shot snippets alone.

 Table~\ref{tab:task_overview} provides illustrative mini-cases for each evaluation dimensions.

\subsection{Task Formulation}

We pose a streaming multi-group protocol to mirror deployment: 5 projects (diverse domains) run independently; each contains $N$ groups that converse daily over a simulated year. This protocol forces systems to make deployment-style decisions—what to store, when to update, and how to attribute—rather than relying on retrospective full-history inspection. During the \textbf{ingestion phase} the system receives daily batches $\{M_{d,g}\}_{g=1}^N$ (chronologically ordered multi-party messages per group) and must autonomously construct and maintain a memory state. During the \textbf{evaluation phase} we pose (1) \textit{multiple-choice} queries for high-precision diagnosis and (2) \textit{open-ended} queries judged by an LLM for semantic equivalence; every query is annotated with evidence spans to enable oracle vs.\ retrieval analyses. This formulation intentionally separates the \emph{storage} problem from the \emph{reasoning} problem so that we can determine whether failures arise from retrieval, representation, or the answer model itself.

\subsection{Data Construction}

Building coherent, long-horizon, multi-party dialogues presents three practical challenges: context truncation, logical drift, and persona/temporal incoherence. We address these not as engineering trivia but as necessary controls: without them, benchmark items risk being unsound (unanswerable) or trivial (solvable without context). Our pipeline (Figure~\ref{curation}) therefore balances realism with repeatability through three stages: (1) blueprint and profile generation, (2) chunk-wise dialogue synthesis with hierarchical summarization, and (3) evidence-grounded QA construction. All generative steps use Gemini-2.5-Pro~\citep{comanici2025gemini25pushingfrontier} for consistency; human-in-the-loop checks ensure plausibility and traceability.

\subsubsection{\textbf{Preliminaries}}
We instantiate a controlled organizational skeleton $S$ with 170 employees $\mathcal{E}$ across 7 departments and five projects $\mathcal{P}$. Each employee receives a persona $\pi_e=(\mathrm{rank}_e,\mathrm{dept}_e,\mathrm{role}_e,\mathbf{s}_e,\mathbf{c}_e)$ where $\mathbf{s}_e$ (40--60 skills) and $\mathbf{c}_e$ (8D style) capture capabilities and communication tendencies. This design choice is essential: roles and skill overlaps create the cross-group dependencies and style shifts that reveal whether a system truly models users rather than matching surface patterns.

\subsubsection{\textbf{Blueprint Generation}}
For each project $p\in\mathcal{P}$ we build a blueprint $B_p=(\mathcal{E}_p,\{(\mathcal{E}_{p,j},\mathcal{T}_{p,j})\}_{j=1}^3)$ that encodes team membership, overlapping assignments, and sub-task timelines. The blueprint enforces global consistency (who can decide what, expected task durations, dependency structure), which is crucial: without a global plan, dialogue synthesis either degenerates into incoherent chatter or becomes artificially easy by exposing explicit decision summaries. The blueprint therefore creates \emph{structured difficulty}—complex but verifiable interactions that mirror real projects.

\subsubsection{\textbf{Conversation Generation}}
Dialogues are generated day-by-day for $D$ simulated days, with each sub-project maintaining a group chat. To preserve long-range coherence under context limits~\citep{wang2025recursively}, we apply hierarchical summarization:
\begin{align}
W_{p,j}^{(d)} &= \mathrm{Summarize}\big(C_{p,j}^{(d-7:d-1)},\; \mathcal{T}_{p,j}^{(d-7:d-1)}\big),\\
M_{p,j}^{(d)} &= \mathrm{Summarize}\big(C_{p,j}^{(d-30:d-1)},\; \mathcal{T}_{p,j}^{(d-30:d-1)}\big),\\
L_{p,j}^{(d)} &= \mathrm{Extract}_{\text{leader}}\big(C_{p,j}^{(d-1)}\big).
\end{align}
These summaries are used exclusively as internal scaffolding during data generation and are never exposed to models during ingestion or evaluation.  Weekly and monthly summaries plus leader instructions compress history while preserving the temporal scaffolding needed to test version reasoning; this is an essential (not cosmetic) mechanism because many temporal errors arise from lost structural cues rather than missing tokens. We generate each day's conversation in a single pass conditioned on these summaries and persona profiles:
\[
C_{p,j}^{(d)} = \mathrm{LLM}_{\text{dialog}}\big(\mathcal{T}_{p,j}^{(d)}, W_{p,j}^{(d)}, M_{p,j}^{(d)}, L_{p,j}^{(d)}, \{\pi_e\}_{e\in\mathcal{E}_{p,j}}\big).
\]
Every generated block is validated by logic, profile, and progress checks; failures trigger bounded regeneration. This produce–verify loop is deliberate: it trades raw spontaneity for reproducible realism so that QA items have precise evidence anchors.

\subsubsection{\textbf{Q\&A Generation}}\label{sec:qa_gen}
We synthesize QAE triples $(q,a,e)$ via three specialized pipelines aligned to the evaluation dimensions. For \textbf{Fine-grained Recall} we use \emph{structure mining} to extract natural multi-hop chains and \emph{non-conflicting implantation} to inject controlled corner cases. For \textbf{Memory Awareness} we construct scenarios of \textit{Updating}, implicit \textit{Constraint} application, and \textit{Proactivity}, and we apply adversarial perturbations (keyword substitution, parameter removal) to force semantic—rather than lexical—retrieval. For \textbf{Profile Understanding} we design distractors that disentangle factual correctness from style matching (a $2\times 2$ fact-vs-style matrix) and craft plausible-but-wrong role inferences to test expertise reasoning.

\subsubsection{\textbf{Q\&A Quality Control}}\label{sec:quality_control}
Quality control is not an afterthought but a methodological core: items that are solvable without context or ambiguous with evidence would nullify diagnostic value. We therefore apply a three-phase filter. \textbf{Phase I (Blind Test)} removes parametric leaks and trivial distractors. \textbf{Phase II (Evidence Grounding)} partitions $\mathcal{C}_p$ into segments $\mathbf{S}$ and enforces \emph{sufficiency} (answer derivable from $S^+$) and \emph{uniqueness} (answer not derivable from any $S^-\,$). \textbf{Phase III (Human Audit)} catches residual logical or pragmatic issues. This pipeline deliberately favors conservative retention: we keep only items with crisp evidence-to-question mappings so that failures can be attributed to memory/ retrieval/ reasoning rather than annotation noise.

These steps also guard against generator-family bias: blueprint specifications fully constrain the factual content of each dialogue, the generator contributes only surface realization; the blind test then ensures that no item whose answer can be inferred from surface.

\subsection{Data Statistics}

\begin{table}[t]
\centering
\small
\setlength{\tabcolsep}{3.5pt}
\begin{tabular}{l r}
\toprule
\textbf{Statistic} & \textbf{Value} \\
\midrule
\multicolumn{2}{l}{\textit{Organizational Structure}} \\
\quad \# Projects / Sub-projects & 5 / 15 \\
\quad \# Employees & 170 \\
\quad \quad Executive / Manager / Staff & 1 / 5 / 164 \\
\quad Avg \# Participants / Project & 37.6 \\
\midrule
\multicolumn{2}{l}{\textit{Dialogue Statistics}} \\
\quad Time Span (days) & 365 \\
\quad Total Dialogue Turns & 51,023 \\
\quad Total Tokens & 4,225,555 \\
\quad Avg Tokens / Project & 845,111 \\
\quad Avg Turns / Day & 28.0 \\
\quad Avg Tokens / Turn & 82.8 \\
\midrule
\multicolumn{2}{l}{\textit{Evaluation Statistics}} \\
\quad \# QA Pairs & 2,400 \\
\quad \# Evaluation Dimensions & 3 \\
\quad \# Sub-dimensions & 9 \\
\bottomrule
\end{tabular}
\caption{\textbf{Data statistics of EverMemBench.}}
\label{tab:data_stats}
\vspace{-13pt}
\end{table}

As summarized in Table~\ref{tab:data_stats}, EverMemBench focuses on dense, deployment-relevant information: 5 projects (Technology, Operations, Marketing, Financial Services, Governance), 170 employees, 51{,}023 turns, and 4.2M tokens (about 1M tokens per project). Unlike distractor-padded long contexts~\citep{wu2025longmemevalbenchmarkingchatassistants}, each 1M-token project contains over 10{,}000 turns of eventful dialogue where a large fraction of events are targeted by QA items. This information density is intentional: it makes retrieval fidelity, attribution, and temporal revision the limiting factors for system performance rather than raw context size.

Figure~\ref{fig:qa_distribution} shows the distribution of 2{,}400 QA pairs across the three dimensions and nine sub-tasks, with balanced coverage across projects (467--493 per topic). These statistics support controlled ablations (e.g., oracle vs.\ retrieved evidence, single- vs.\ multi-group questions) that reveal whether errors stem from retrieval granularity, evidence fragmentation, or the answer model's reasoning capability.

\begin{figure}[t]
\centering
\includegraphics[width=.8\linewidth]{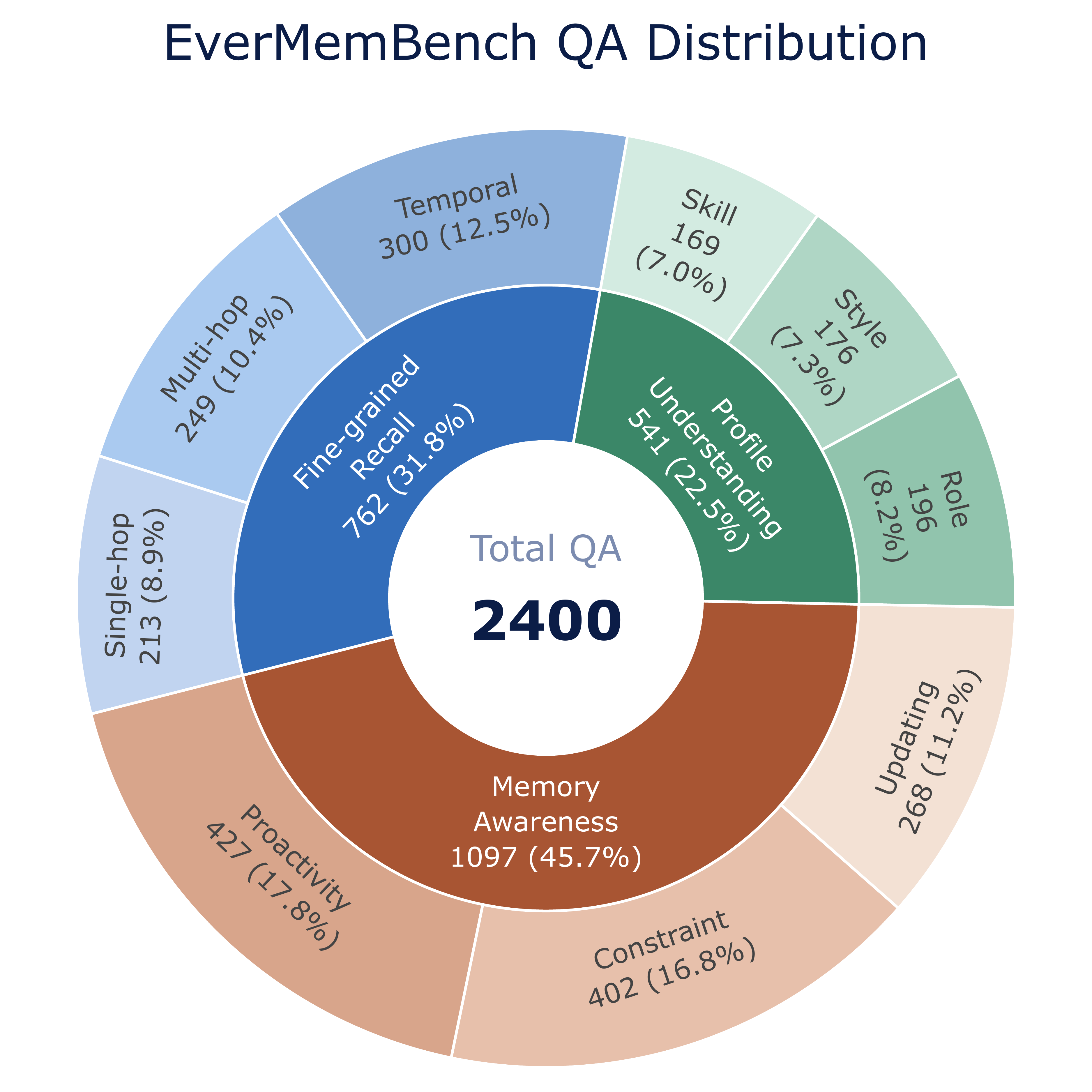}
\caption{Distribution of QA pairs.}
\label{fig:qa_distribution}
\end{figure}

\begin{table*}[t]
 \centering
\fontsize{8pt}{8pt}\selectfont
\setlength{\tabcolsep}{3.5pt}
 \renewcommand{\arraystretch}{1.15}
 \newcommand{\ci}[1]{{\tiny\textcolor{black}{$\pm$#1}}}

 \resizebox{0.85\textwidth}{!}{%
 \begin{tabular}{l *{3}{c} *{3}{c} *{3}{c} c}
 \toprule
 \multirow{2}{*}{\textbf{Method}}
 & \multicolumn{3}{c}{\textbf{Fine-Grained Recall}}
 & \multicolumn{3}{c}{\textbf{Memory Awareness}}
 & \multicolumn{3}{c}{\textbf{Profile Understanding}}
 & \multirow{2}{*}{\textbf{Average}} \\
 \cmidrule(lr){2-4}\cmidrule(lr){5-7}\cmidrule(lr){8-10}
 & \textit{Single} & \textit{Multi} & \textit{Temp}
 & \textit{Const} & \textit{Proact} & \textit{Update}
 & \textit{Style} & \textit{Skill} & \textit{Role} & \\
 \midrule
 
 \multicolumn{11}{c}{\cellcolor{gray!10}\textbf{GPT-4.1-mini}} \\
 \midrule
 Full Context & 83.57\ci{4.9} & 2.41\ci{1.8} & 7.00\ci{2.8} & 63.43\ci{4.7} & 25.06\ci{4.1} & 42.54\ci{6.0} & 39.20\ci{7.4} & 35.50\ci{7.1} &
 38.27\ci{6.9} & 37.44\ci{1.8} \\
 \cdashline{1-11}
 \addlinespace[2pt]
 \hspace{1em} + MemoBase & 60.09\ci{6.6} & 12.85\ci{4.2} & \textbf{18.00}\ci{4.3} & 64.68\ci{4.6} & 36.77\ci{4.6} & 30.60\ci{5.6} & 17.05\ci{5.4} &
 29.59\ci{6.8} & 38.78\ci{6.9} & 34.27\ci{1.9} {\scriptsize\textcolor{red}{(-3.18)}} \\
 \hspace{1em} + Mem0 & 55.40\ci{6.6} & 11.24\ci{3.8} & 6.33\ci{2.8} & 66.17\ci{4.6} & \textbf{52.46}\ci{4.7} & \textbf{51.87}\ci{6.0} & 22.73\ci{6.3} &
 31.36\ci{7.1} & 36.22\ci{6.9} & 37.09\ci{1.9} {\scriptsize\textcolor{red}{(-0.36)}} \\
 \hspace{1em} + Zep & \textbf{73.71}\ci{5.9} & 8.03\ci{3.4} & 13.00\ci{3.8} & 67.16\ci{4.6} & 47.54\ci{4.7} & 43.66\ci{6.0} & 26.70\ci{6.5} & \textbf{35.50}\ci{7.1} &
  44.39\ci{6.9} & 39.97\ci{1.9} {\scriptsize\textcolor{teal}{(+2.52)}} \\
 \hspace{1em} + MemOS & 71.36\ci{6.1} & \textbf{18.88}\ci{4.8} & 15.67\ci{4.0} & \textbf{69.90}\ci{4.5} & 51.99\ci{4.7} & 45.15\ci{6.0} & \textbf{28.98}\ci{6.5} &
 32.54\ci{7.1} & \textbf{48.47}\ci{7.1} & \textbf{42.55}\ci{1.9} {\scriptsize\textcolor{teal}{(+5.11)}} \\

 \midrule
 \multicolumn{11}{c}{\cellcolor{gray!10}\textbf{Llama-4-Scout-17B-16E-Instruct}} \\
 \midrule
 Full Context & 77.93\ci{5.6} & 0.00\ci{0.0} & 1.67\ci{1.5} & 60.45\ci{4.7} & 43.79\ci{4.7} & 67.91\ci{5.6} & 27.84\ci{6.8} & 39.64\ci{7.4} &
 42.35\ci{6.9} & 40.18\ci{1.8} \\
 \cdashline{1-11}
 \addlinespace[2pt]
 \hspace{1em} + MemoBase & 57.75\ci{6.6} & 5.62\ci{3.0} & \textbf{12.00}\ci{3.8} & \textbf{67.41}\ci{4.6} & 54.10\ci{4.7} & 27.61\ci{5.4} &
 21.02\ci{6.0} & 47.34\ci{7.7} & 42.86\ci{6.6} & 37.30\ci{1.8} {\scriptsize\textcolor{red}{(-2.88)}} \\
 \hspace{1em} + Mem0 & 56.34\ci{6.6} & 3.21\ci{2.2} & 3.67\ci{2.2} & 66.17\ci{4.5} & 63.00\ci{4.7} & \textbf{45.90}\ci{6.0} & 23.30\ci{6.0} &
 51.48\ci{7.7} & 44.39\ci{6.9} & 39.72\ci{1.8} {\scriptsize\textcolor{red}{(-0.46)}} \\
 \hspace{1em} + Zep & \textbf{71.36}\ci{6.1} & 4.02\ci{2.4} & 7.00\ci{2.8} & \textbf{67.41}\ci{4.5} & 52.69\ci{4.7} & 35.45\ci{6.0} & \textbf{27.84}\ci{6.8} &
 46.15\ci{7.7} & 46.43\ci{7.1} & 39.82\ci{1.8} {\scriptsize\textcolor{red}{(-0.36)}} \\
 \hspace{1em} + MemOS & 67.61\ci{6.1} & \textbf{6.43}\ci{3.0} & 11.33\ci{3.5} & 66.92\ci{4.5} & \textbf{64.17}\ci{4.7} & 38.43\ci{6.0} & 23.86\ci{6.2} &
 \textbf{53.25}\ci{7.7} & \textbf{50.00}\ci{7.1} & \textbf{42.44}\ci{1.9} {\scriptsize\textcolor{teal}{(+2.27)}} \\

 \midrule
 \multicolumn{11}{c}{\cellcolor{gray!10}\textbf{Gemini-3-Flash}} \\
 \midrule
 Full Context & 97.65\ci{2.1} & 26.51\ci{5.4} & 45.00\ci{5.7} & 96.77\ci{1.7} & 98.36\ci{1.2} & 100.00\ci{0.0} & 67.05\ci{6.8} & 53.25\ci{7.7} &
 68.88\ci{6.6} & 72.61\ci{1.6} \\
 \cdashline{1-11}
 \addlinespace[2pt]
 \hspace{1em} + MemoBase & 56.34\ci{6.6} & 6.43\ci{3.0} & 17.67\ci{4.2} & 85.32\ci{3.5} & \textbf{91.10}\ci{2.6} & 84.33\ci{4.3} & 38.07\ci{7.4} &
 53.85\ci{7.7} & 69.39\ci{6.6} & 55.83\ci{1.8} {\scriptsize\textcolor{red}{(-16.78)}} \\
 \hspace{1em} + Mem0 & 56.34\ci{6.6} & 5.62\ci{3.0} & 2.67\ci{1.8} & 79.60\ci{3.9} & 84.54\ci{3.4} & 85.45\ci{4.3} & 36.93\ci{7.1} & 56.21\ci{7.4} &
  61.73\ci{6.6} & 52.12\ci{1.8} {\scriptsize\textcolor{red}{(-20.48)}} \\
 \hspace{1em} + Zep & 68.54\ci{6.1} & 6.02\ci{3.0} & 11.00\ci{3.5} & \textbf{85.82}\ci{3.4} & 82.44\ci{3.5} & 78.36\ci{4.9} & 34.66\ci{6.8} &
 60.95\ci{7.4} & 66.33\ci{6.6} & 54.90\ci{1.8} {\scriptsize\textcolor{red}{(-17.71)}} \\
 \hspace{1em} + MemOS & \textbf{69.01}\ci{6.1} & \textbf{10.84}\ci{3.8} & \textbf{20.67}\ci{4.7} & 81.84\ci{3.7} & 87.59\ci{3.0} & \textbf{90.67}\ci{3.5} &
 \textbf{38.64}\ci{7.1} & \textbf{62.72}\ci{7.1} & \textbf{71.43}\ci{6.4} & \textbf{59.27}\ci{1.8} {\scriptsize\textcolor{red}{(-13.34)}} \\

 \bottomrule
 \end{tabular}%
 }

   \caption{Main evaluation results on EverMemBench. "Full Context" uses the complete dialogue history; memory-augmented methods use only retrieved information. Best memory-augmented results per metric are bolded. Parenthesized values show accuracy change vs.\ Full Context. Gray subscripts are half-widths of 95\% bootstrap CIs ($B{=}10{,}000$).}
 \label{tab:main_results}
 \vspace{-13pt}
 \end{table*}

\section{Empirical Results}

\nop{
We evaluate LLMs and memory-augmented systems on EverMemBench to diagnose how they handle multi-party, high-density conversational dynamics.

\subsection{Experimental Setup}

\paragraph{Evaluated Systems.}
We evaluate two categories: (1) \textit{Long-context LLMs} that receive complete conversation history, including commercial models Gemini-3-Flash~\citep{google2025gemini3flash} and GPT-4.1-mini~\citep{openai2024gpt4}, as well as the open-source LLaMA-4-Scout~\citep{meta2025llama4}, all supporting 1M tokens context windows; (2) \textit{Memory-augmented systems}, including Zep~\citep{rasmussen2025zeptemporalknowledgegraph}, Mem0~\citep{chhikara2025mem0buildingproductionreadyai}, MemOS~\citep{li2025memosoperatingmemoryaugmentedgeneration} and MemoBase~\citep{memobase2025}. We access all memory systems through their official cloud APIs. Each system is evaluated using its official retrieval configuration reported for LoCoMo: Zep and Mem0 retrieve top-$k$=10 memories; MemOS retrieves top-$k$=20; MemoBase retrieves up to 3K tokens per query. All LLM inference uses greedy decoding (temperature\,=\,0).

\paragraph{Oracle Evaluation.}
To establish upper-bound performance, we construct an \textit{oracle} setting where ground-truth evidence spans are provided directly to the LLM. For each question, we extract the minimal set of conversation segments that contain the answer, bypassing the retrieval process entirely. This isolates the LLM's reasoning capability from retrieval quality, revealing whether performance gaps stem from retrieval failures or fundamental reasoning limitations. Profile Understanding is excluded from this evaluation since user traits like communication style and expertise are never stated explicitly but emerge from cumulative interaction patterns, leaving no discrete evidence spans that could serve as oracle input.

\begin{table}[t]
\centering
\small
\fontsize{8pt}{10pt}\selectfont
\setlength{\tabcolsep}{3.5pt}
\renewcommand{\arraystretch}{1.1}

\begin{tabular}{l ccc ccc}
\toprule
\multirow{2}{*}{\textbf{Model}}
& \multicolumn{3}{c}{\textbf{Fine-Grained Recall}}
& \multicolumn{3}{c}{\textbf{Memory Awareness}} \\
\cmidrule(lr){2-4}\cmidrule(lr){5-7}
& \textit{Single} & \textit{Multi} & \textit{Temp}
& \textit{Const} & \textit{Proact} & \textit{Update} \\
\midrule
GPT-4.1-mini  & 99.53 & 97.99 & 60.00 & 96.77 & 86.65 & 98.51  \\
Llama-4-Scout & 96.24 & 37.35 & 34.00 & 93.53 & 90.87 & 96.64  \\
Gemini-3-Flash & 98.14 & 88.37 & 54.33 & 99.26 & 98.12 & 99.23  \\
\bottomrule
\end{tabular}
\caption{Oracle performance with ground-truth evidence spans provided directly to the model, isolating reasoning capability from retrieval quality.}
\label{tab:oracle}
\vspace{-15pt}
\end{table}

\paragraph{Evaluation Metrics.}
For \textit{Fine-Grained Recall}, answers are concrete facts (names, numbers, timestamps) that may vary in surface form while remaining semantically correct; we use LLM-as-a-judge~\citep{zheng2023judgingllmasajudgemtbenchchatbot} with Gemini-3-Flash as the judge model to assess semantic equivalence. For \textit{Memory Awareness} and \textit{Profile Understanding}, the challenge lies in distinguishing genuine memory recall from plausible fabrication---a system without true memory access could still generate reasonable-sounding responses by leveraging general knowledge. To isolate actual recall ability, we use multiple-choice format with carefully constructed distractors that are contextually plausible but factually inconsistent with the conversation history.
\begin{figure}[t]
	\begin{center}
        \includegraphics[width=\linewidth]{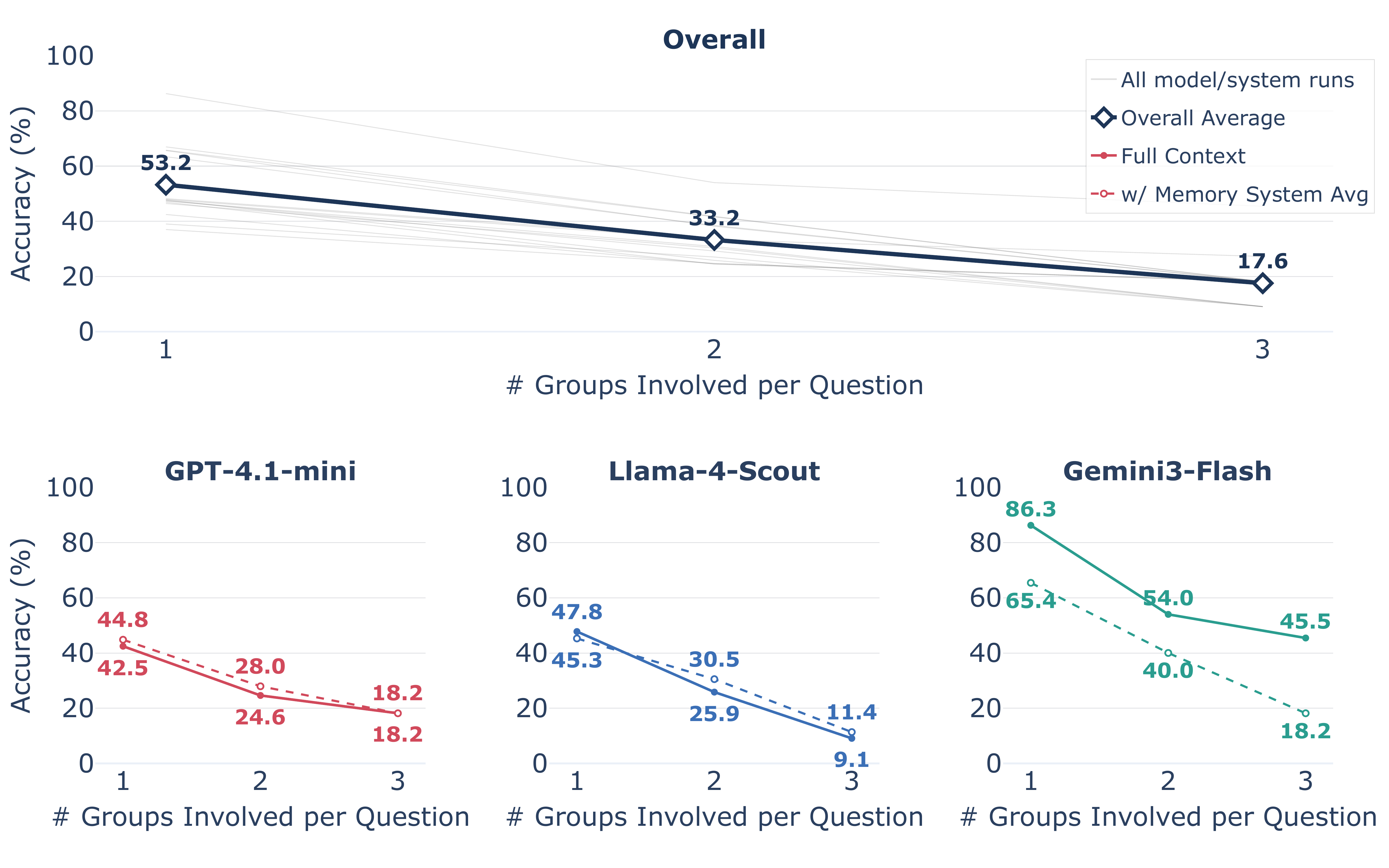}
	\end{center}
    \vspace{-13pt}
    	\caption{Accuracy by number of groups involved per question. Top: overall average across all settings. Bottom: breakdown by answer model.}
	\label{fig:acc_by_groups}
\end{figure}

\subsection{Fine-Grained Recall Analysis.}
\textbf{Multi-hop and temporal reasoning collapse under multi-party interleaving.} Multi-hop and temporal questions expose fundamental limitations of both long-context and retrieval-based approaches.

All systems achieve reasonable single-hop accuracy (Gemini-3-Flash: 97.65\%, memory systems: 55--83\%), yet multi-hop questions cause universal performance collapse. Gemini-3-Flash drops from 97.65\% to 26.51\%, and the best memory system reaches only 18.88\%. This sharp contrast reveals that the challenge lies not in accessing individual facts, but in integrating information scattered across speakers, groups, and time points. A decision may be proposed by one person, debated in another channel, and finalized by a third party days later. Neither brute-force context access nor retrieval-based memory handles this cross-source integration adequately.

Oracle evaluation confirms that this is primarily a retrieval problem for strong models: when ground-truth evidence spans are provided directly, GPT-4.1-mini improves from 2.41\% to 97.99\% and Gemini-3-Flash from 26.51\% to 88.37\% (Table~\ref{tab:oracle}). However, LLaMA-4 achieves only 37.35\% even with oracle evidence, as it frequently refuses to answer or hedges when presented with fragmented dialogue segments, interpreting them as incomplete information.

This fragmentation intensifies as the number of groups involved increases. As shown in Figure~\ref{fig:cross_group}, accuracy drops sharply from 54.5\% (single-group) to 33.6\% (two-group) to 19.7\% (three-group), representing a 64\% relative decline. The degradation is universal across all models and configurations, confirming that cross-group information integration remains a fundamental bottleneck.

Temporal questions yield uniformly poor results: Gemini-3-Flash achieves only 45.00\%, GPT-4.1-mini scores 7.00\%, and memory systems range from 2.67\% to 21.00\%. In realistic collaboration, decisions evolve over time. Answering ``What is the final budget?'' requires not just locating mentions but understanding version semantics: which statement supersedes which, and what constitutes the final state. Current architectures treat memory as a flat store rather than a versioned timeline, leaving temporal reasoning largely unsolved. Notably, this limitation persists even under oracle conditions: the best model (GPT-4.1-mini) achieves only 60\%, while LLaMA-4 reaches just 34\% (Table~\ref{tab:oracle}), indicating a fundamental reasoning bottleneck beyond retrieval.

\subsection{Memory Awareness Analysis.}
\textbf{Retrieval fails to capture inferential relevance.} Oracle evaluation establishes Memory Awareness as a retrieval-bound problem: with ground-truth evidence, all models achieve 87--99\% (Table~\ref{tab:oracle}), confirming sufficient reasoning capability. The bottleneck lies in surfacing the right information.

Under the full-context setting, Gemini-3-Flash maintains 97--100\%, approaching oracle performance, while GPT-4.1-mini drops to 25--63\% and LLaMA-4 to 43--68\%. This reveals that weaker models struggle to locate relevant information within lengthy multi-party dialogues, whereas Gemini's superior long-context reasoning enables effective evidence identification.

Under the memory-augmented setting, retrieval helps weaker models by filtering noise (GPT-4.1-mini improves from 25--63\% to 29--76\%), but degrades Gemini's performance (from 97--100\% to 76--90\%) by discarding contextual cues. Crucially, even with identical retrieved evidence, model capability still creates substantial gaps: Gemini achieves 76--90\% while GPT-4.1-mini and LLaMA-4 score only 29--76\% and 35--73\% respectively, demonstrating that stronger models better exploit weak evidence, i.e., fragments lacking explicit connections, by inferring answers through elimination of implausible options.

These findings reveal a dual bottleneck: (1) retrieval systems fail to capture inferential relevance, such as connecting a restaurant query to a colleague's dietary restriction mentioned weeks prior; and (2) answer models must reason robustly over sparse, indirect cues when retrieval provides only partial information. The gap between retrieval relevance and reasoning relevance represents a fundamental challenge for current memory architectures.

\subsection{Profile Understanding Analysis.}
The difficulty is fundamental: communication style is not a discrete fact but a pattern that emerges from many interactions. Whether a user is directive versus collaborative, formal versus casual, cannot be captured by any single snippet. It requires aggregating signals distributed across the entire dialogue history. Current memory paradigms, designed to retrieve discrete facts, are structurally ill-suited for this kind of pattern abstraction. In contrast, Skill and Title are more amenable to extraction because they can be partially inferred from individual task discussions or organizational context, even though performance on these subtasks also remains far from solved.

}

We evaluate LLMs and memory-augmented systems on EverMemBench to understand \emph{why} current approaches fail under realistic collaborative interaction, not merely \emph{how much} they fail. The experiments are designed to disentangle three factors that are often conflated in prior work: (i) access to long context, (ii) retrieval quality, and (iii) reasoning over fragmented, evolving evidence. By systematically varying evidence access (full context, retrieval, oracle), we show that many failures observed in practice are structural and anticipated consequences of multi-party, multi-group interaction.

\subsection{Experimental Setup}

\paragraph{Evaluated Systems.}
We evaluate two categories of systems. \textit{Long-context LLMs} consume the complete dialogue history, including Gemini-3-Flash~\citep{google2025gemini3flash}, GPT-4.1-mini~\citep{openai2024gpt4}, and LLaMA-4-Scout~\citep{meta2025llama4}, all supporting 1M-token contexts. \textit{Memory-augmented systems} attach external memory to an answer model via similarity-based retrieval, including Zep~\citep{rasmussen2025zeptemporalknowledgegraph}, Mem0~\citep{chhikara2025mem0buildingproductionreadyai}, MemOS~\citep{li2025memosoperatingmemoryaugmentedgeneration} and MemoBase~\citep{memobase2025}. We use official cloud APIs and default retrieval configurations reported for LoCoMo: Zep and Mem0 retrieve top-$k{=}10$ items; MemOS retrieves top-$k{=}20$; MemoBase retrieves up to 3K tokens. Under this configuration, memory-augmented systems consume roughly 1K--3K input tokens per query, whereas full-context baselines ingest the entire ${\sim}$1M-token dialogue history. Unless noted otherwise, all memory-augmented systems use GPT-4.1-mini as the answer model, allowing us to isolate retrieval effects. 

\paragraph{Oracle Evaluation.}
To separate retrieval failures from reasoning limitations, we construct an oracle setting in which the ground-truth dialogue segments from which each answer must be synthesized are provided directly to the LLM. By bypassing retrieval entirely, oracle evaluation isolates each sub-dim's reasoning demands over the provided evidence, which range from entity matching for Single-hop to lifecycle boundary disambiguation and date arithmetic for Temporal (Figure~\ref{fig:recall_cases_part3}). Table 5 reports oracle performance. Profile Understanding is excluded from this setting because persona traits are implicit and distributed, leaving no discrete evidence spans that could serve as oracle input. Oracle performance thus defines an upper bound for any retrieval-only improvement, as it reflects model behavior when all relevant evidence is perfectly surfaced.

\begin{table}[t]
\centering
\small
\fontsize{8pt}{10pt}\selectfont
\setlength{\tabcolsep}{3.5pt}
\renewcommand{\arraystretch}{1.1}

\begin{tabular}{l ccc ccc}
\toprule
\multirow{2}{*}{\textbf{Model}}
& \multicolumn{3}{c}{\textbf{Fine-Grained Recall}}
& \multicolumn{3}{c}{\textbf{Memory Awareness}} \\
\cmidrule(lr){2-4}\cmidrule(lr){5-7}
& \textit{Single} & \textit{Multi} & \textit{Temp}
& \textit{Const} & \textit{Proact} & \textit{Update} \\
\midrule
GPT-4.1-mini  & 99.53 & 97.99 & 60.00 & 96.77 & 86.65 & 98.51  \\
Llama-4-Scout & 96.24 & 37.35 & 34.00 & 93.53 & 90.87 & 96.64  \\
Gemini-3-Flash & 98.14 & 88.37 & 54.33 & 99.26 & 98.12 & 99.23  \\
\bottomrule
\end{tabular}
\caption{Oracle performance with ground-truth evidence spans provided directly to the model, isolating reasoning capability from retrieval quality.}
\label{tab:oracle}
\vspace{-15pt}
\end{table}

\paragraph{Evaluation Metrics.}
For \textit{Fine-Grained Recall}, answers are concrete facts (names, numbers, timestamps) that may vary lexically while remaining semantically correct; we use LLM-as-a-judge~\citep{zheng2023judgingllmasajudgemtbenchchatbot} to assess equivalence. For \textit{Memory Awareness} and \textit{Profile Understanding}, we use multiple-choice questions with carefully constructed distractors to prevent plausible fabrication by models without true memory access. This design ensures that high scores reflect genuine memory use rather than surface-level plausibility.

\subsection{Fine-Grained Recall: Attribution and Time as Structural Bottlenecks}

\textbf{Multi-hop and temporal reasoning collapse under multi-party interleaving.}
As summarized in Table~\ref{tab:main_results}, all systems perform well on \textit{Single-hop} recall (Gemini-3-Flash: 97.65\%; memory systems: 55--83\%), indicating that isolated fact retrieval is largely solved. However, \textit{Multi-hop} accuracy drops sharply: Gemini-3-Flash falls to 26.51\%, and the best memory-augmented system reaches only 18.88\%. This is not a recall problem but an integration problem. In EverMemBench, relevant facts are distributed across speakers, groups, and days; answering correctly requires stitching together partial evidence that never co-occurs in a single exchange.

Oracle results in Table~\ref{tab:oracle} confirm this diagnosis. When provided with ground-truth evidence, GPT-4.1-mini improves from 2.41\% to 97.99\% and Gemini-3-Flash from 26.51\% to 88.37\%, showing that strong models can reason correctly once attribution and retrieval are removed as bottlenecks. In contrast, LLaMA-4 reaches only 37.35\% even under oracle conditions, frequently refusing to answer when evidence appears fragmented—highlighting a distinct failure mode rooted in conservative reasoning rather than retrieval.

The difficulty compounds as information spans more groups. Figure~\ref{fig:acc_by_groups} shows that accuracy drops from 54.5\% (single-group) to 33.6\% (two groups) and 19.7\% (three groups), a 64\% relative decline. This degradation is consistent across models and memory configurations, demonstrating that cross-group attribution—not context length—is the dominant challenge.

\begin{figure}[t]
	\begin{center}
        \includegraphics[width=\linewidth]{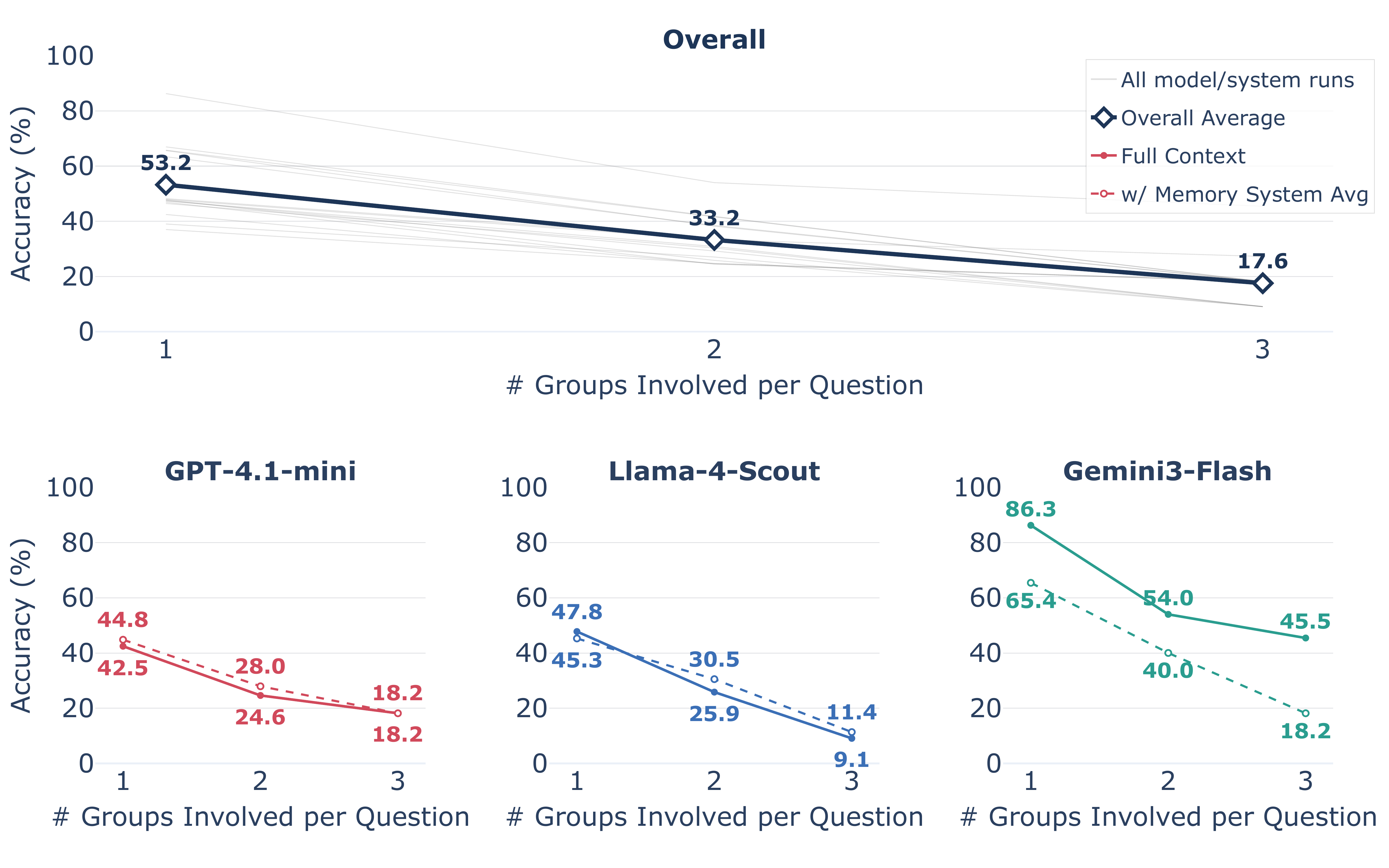}
	\end{center}
    \vspace{-13pt}
    	\caption{Accuracy by number of groups involved per question. Top: overall average across all settings. Bottom: breakdown by answer model.}
	\label{fig:acc_by_groups}
\end{figure}

\textbf{Temporal reasoning remains unsolved.}
Temporal questions expose a complementary limitation. Across all systems, performance is low (Gemini-3-Flash: 45.00\%; GPT-4.1-mini: 7.00\%; memory systems: 2.67--21.00\%). In realistic collaboration, decisions are revised, superseded, and finalized over time, and answering correctly requires reasoning over version semantics, not just timestamps. This difficulty persists even with oracle evidence: the best model reaches only 60\% and LLaMA-4 only 34\% (Table~\ref{tab:oracle}), because the provided evidence necessarily contains overlapping lifecycle signals such as premature completion announcements and archival statements from other speakers on adjacent dates, which the model must still disambiguate before computing durations. These results indicate a reasoning gap that current memory architectures cannot bridge, as they treat time as an ordering signal rather than a semantic construct that encodes event lifecycle stages such as initiation, revision, and completion.

\subsection{Memory Awareness: Retrieval vs.\ Reasoning}

\textbf{Retrieval misses inferentially relevant evidence.}
Memory Awareness tasks are explicitly designed to test whether systems recognize when past information matters. Oracle results in Table~\ref{tab:oracle} show that all models achieve 87--99\% accuracy, confirming that reasoning capability is sufficient when relevant evidence is available. The difficulty lies in retrieving \emph{the right} evidence.

Under full-context access, Gemini-3-Flash maintains near-oracle performance (97--100\%), while GPT-4.1-mini and LLaMA-4 degrade substantially (25--63\% and 43--68\%). This gap reflects differences in long-context reasoning: weaker models struggle to locate relevant signals amid dense, multi-party dialogue.

Memory augmentation partially compensates for weaker models by filtering noise: GPT-4.1-mini improves to 29--76\%. However, the same retrieval pipelines degrade Gemini-3-Flash (down to 76--90\%) by discarding contextual cues the model could otherwise exploit. Even with identical retrieved evidence, performance diverges sharply—Gemini reaches 76--90\%, while GPT-4.1-mini and LLaMA-4 reach only 29--76\% and 35--73\%. This reveals a dual bottleneck: retrieval fails to surface \emph{inferentially relevant} information, and weaker answer models struggle to reason over sparse, indirect cues.

\subsection{Profile Understanding: Emergent Patterns Resist Retrieval}

Profile Understanding tasks probe whether systems can maintain consistent user models over time. Results in Table~\ref{tab:main_results} show that \textit{Style} is the most challenging subtask, with memory systems achieving only 11--58\% and even Gemini-3-Flash reaching 67.05\%. Unlike factual recall, communication style is an emergent pattern spanning many interactions; it cannot be recovered from isolated snippets. In contrast, \textit{Skill} and \textit{Role} achieve higher accuracy because they are partially inferable from individual task contexts and organizational structure. These results underscore a limitation of retrieval-centric memory: it excels at discrete facts but struggles with traits that are never stated and can only be captured by aggregating behavioral signals across the full interaction history.

\paragraph{Takeaway.}
Across all analyses, Figures~\ref{fig:acc_by_groups} and Tables~\ref{tab:main_results} and~\ref{tab:oracle} jointly show that scaling context or retrieval alone is insufficient. Multi-party attribution, temporal revision, and inferential relevance introduce structural challenges that current memory paradigms do not address. EverMemBench surfaces these failures by design, providing a diagnostic benchmark that shifts evaluation from leaderboard comparisons to understanding \emph{why} memory systems fail—and what next-generation architectures must change to succeed.

\section{Conclusion}

In this paper, we introduced EverMemBench, a high-realism benchmark for long-term conversational memory that reflects how LLMs are used in practice: as participants in sustained, multi-party collaboration where information is distributed across speakers and groups, revised over time, and implicitly shaped by roles and social context. Through a carefully controlled curation pipeline grounded in project timelines and public events, EverMemBench produces traceable dialogue logs and evidence-grounded QA items that stress memory challenges beyond context length, including attribution, temporal revision, and inferential relevance. Our empirical results show that many failures observed in deployed systems are structural rather than incidental: multi-hop reasoning collapses under multi-party attribution, temporal reasoning fails without explicit version semantics, and similarity-based retrieval struggles to surface implicitly relevant information even when reasoning capacity is sufficient.

These findings suggest clear directions for future research. First, memory architectures must move beyond flat or snippet-based storage toward representations that explicitly encode versioned state, episodic boundaries, and cross-group dependencies. Second, effective memory systems will need to integrate retrieval and reasoning more tightly, enabling models to recognize inferential relevance rather than relying solely on surface similarity. By decomposing memory competence into fine-grained recall, memory awareness, and profile understanding, EverMemBench provides a diagnostic foundation for systematically studying these challenges. We view EverMemBench as a cornerstone benchmark for the next generation of LLMs, one that enables principled progress toward structured, time-aware memory and socially grounded reasoning in realistic collaborative settings.


\bibliographystyle{ACM-Reference-Format}
\bibliography{sample-base}

\appendix
\clearpage

\begin{figure*}[t]
\centering
\includegraphics[width=.8\linewidth]{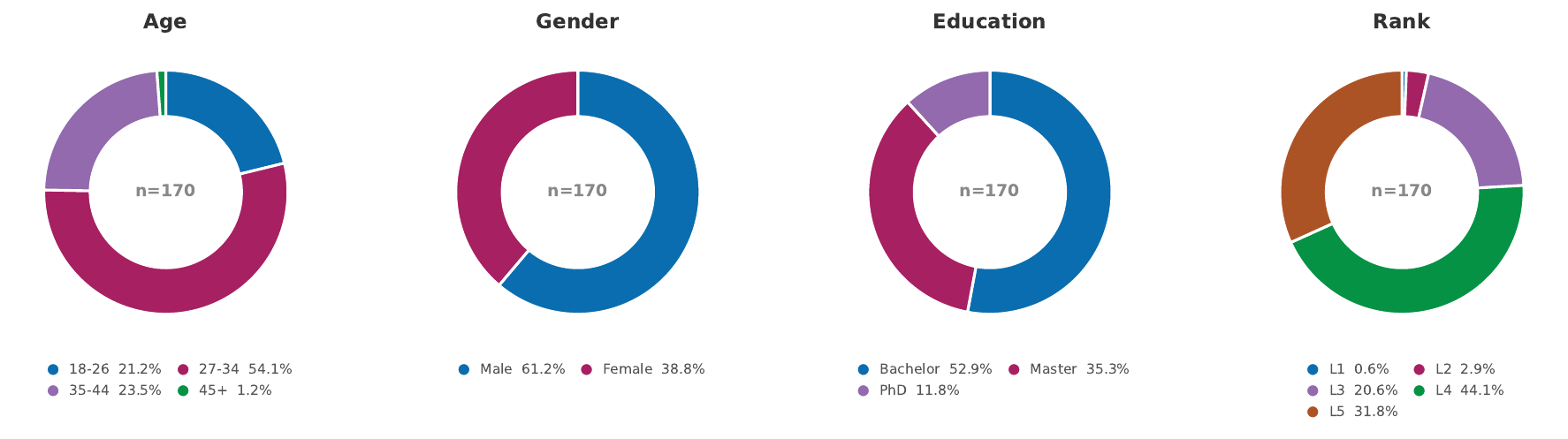}
 \vspace{-13pt}
\caption{Demographic distributions of the 170 simulated user profiles across four dimensions: age, gender, education level, and organizational rank.}
\label{fig:profile_ring}
\end{figure*}
\begin{figure*}[t]
\centering
\includegraphics[width=.8\linewidth]{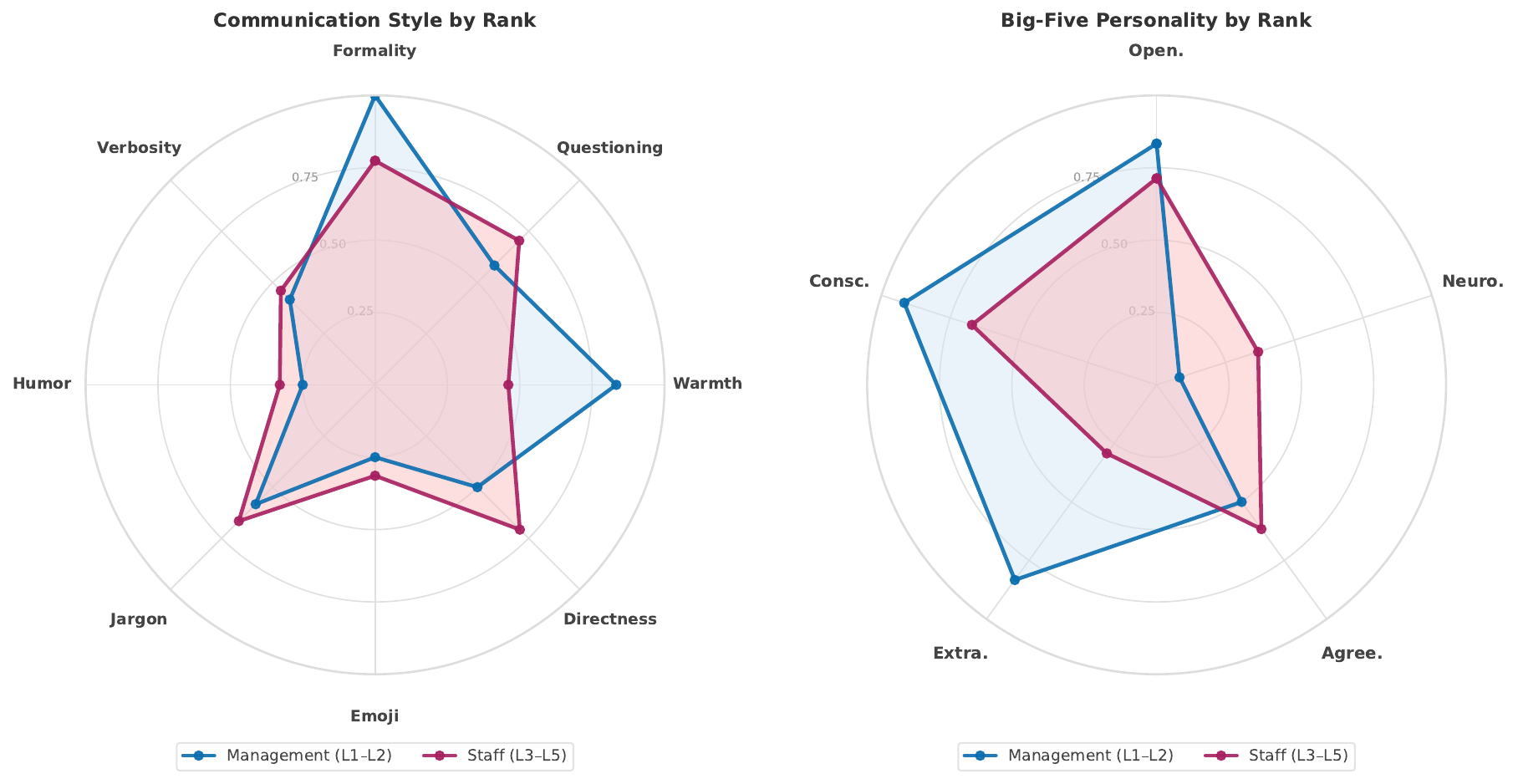}
 \vspace{-10pt}
\caption{Radar charts comparing communication style (left) and Big-Five personality traits (right) between management (L1--L2) and staff (L3--L5) groups.}
\label{fig:profile_rodar}
\end{figure*}
\begin{figure}[t]
\centering
\includegraphics[width=.8\linewidth]{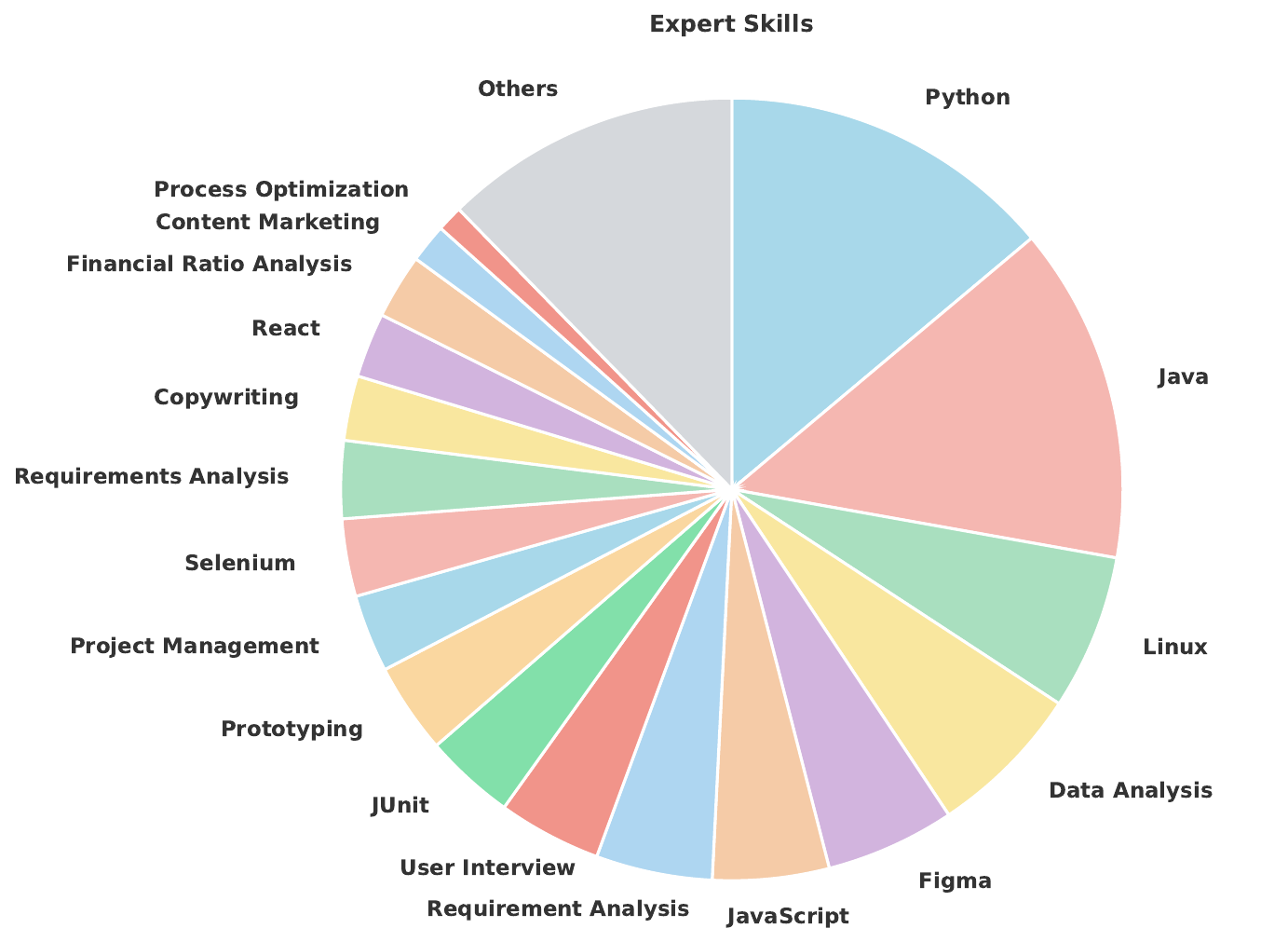}
 \vspace{-10pt}
\caption{Distribution of expert skills among simulated user profiles. The top-18 skills are shown individually; skills representing less than 1\% of the total are       
  grouped into ``Others.''}
\label{fig:profile_skill}
\end{figure}

 \section{Data Statistics}\label{appendix:statistics}                                                                                                        
  \subsection{Profile Distribution}                                                                                                                                    
  Our benchmark comprises \textbf{170} unique employee profiles spanning a simulated enterprise organization. Each profile is characterized by
  demographic attributes, a skill portfolio, an 8-dimensional communication style, and a Big-Five personality trait vector.

  \paragraph{Demographics.}
  Figure~\ref{fig:profile_ring} summarizes the demographic composition. Ages range from 23 to 48 (mean 30.8), with the majority (54.1\%) in the
  27--34 bracket. The gender split is 61.2\% male and 38.8\% female. Education levels include Bachelor's (52.9\%), Master's (35.3\%), and PhD (11.8\%).
  The organizational hierarchy follows a five-level rank system: L1 (CEO, 0.6\%), L2 (Directors, 2.9\%), L3 (Senior Staff, 20.6\%), L4 (Mid-level,
  44.1\%), and L5 (Junior, 31.8\%), reflecting a realistic corporate pyramid.

  \paragraph{Skill Profile.}
  Each employee is assigned skills drawn from a pool of \textbf{104} unique competencies, with an average of 4.1 skills per member (range: 3--7). Skill
  assignments are rank-dependent:
  \begin{itemize}
      \item \textbf{Rank 1--2 (Executives)}: 6--10 skills, including strategic management and cross-functional competencies.
      \item \textbf{Rank 3 (Senior Staff)}: 4--6 skills, combining domain expertise with team coordination.
      \item \textbf{Rank 4--5 (Staff)}: 3--5 skills, focused on role-specific technical or business competencies.
  \end{itemize}
  Each skill is annotated with a proficiency level: \textit{strong} (27.2\%), \textit{medium} (47.9\%), or \textit{low} (24.9\%). As shown in
  Figure~\ref{fig:profile_skill}, the most prevalent expert-level skills are Python and Java (each 13.9\%), followed by Linux and Data Analysis (each
  6.4\%). Skills with less than 1\% share are grouped into ``Others.''

  \paragraph{Communication Style.}
  Each persona is assigned an 8-dimensional communication profile (Figure~\ref{fig:profile_rodar}, left):
  \begin{itemize}
      \item \textbf{Formality}: Semi-formal (78.2\%) / Casual (21.8\%)
      \item \textbf{Verbosity}: Moderate (38.8\%) / Concise (34.7\%) / Detailed (26.5\%)
      \item \textbf{Humor}: Minimal (50.0\%) / Occasional (34.7\%) / Frequent (15.3\%)
      \item \textbf{Jargon Usage}: Technical (50.6\%) / Balanced (30.6\%) / Plain (18.2\%)
      \item \textbf{Emoji Usage}: Rare (54.7\%) / Occasional (28.2\%) / Frequent (17.1\%)
      \item \textbf{Directness}: Balanced (57.6\%) / Direct (41.2\%) / Indirect (1.2\%)
      \item \textbf{Warmth}: Friendly (81.8\%) / Neutral (11.8\%) / Warm (6.5\%)
      \item \textbf{Questioning Style}: Probing (47.6\%) / Clarifying (44.7\%) / Accepting (7.6\%)
  \end{itemize}
  Assignments are conditioned on role expectations: executives tend toward formal, direct, and neutral styles, while technical staff favor concise,
  technical, and minimal-humor communication.

  \paragraph{Personality Traits.}
  Big-Five personality traits are assigned at three levels (High/Medium/Low), as visualized in Figure~\ref{fig:profile_rodar} (right). Openness skews
  high (44.7\% High, 54.1\% Medium), Extraversion skews low (51.8\% Low), and Agreeableness is predominantly medium (77.6\%). Management-level employees
  (L1--L2) exhibit markedly higher Openness and Extraversion compared to staff (L3--L5), consistent with leadership role expectations.

  \paragraph{Overall Analysis}
These distributions are intentionally skewed to mirror realistic enterprise communication norms---semi-formal tone and friendly warmth predominate in professional group chats, while the rank pyramid (75.9\% at L4--L5) reflects typical corporate hierarchies. Although certain individual dimensions are concentrated (e.g., Semi-formal 78.2\%, Friendly 81.8\%), the impact on benchmark validity is limited: each persona is characterized by a \emph{combination} of all eight style dimensions, so even personas sharing the same value on one axis differ substantially along others---the combinatorial space ensures diverse and distinguishable style fingerprints across the 170 employees. Moreover, as shown in Figure~\ref{fig:profile_rodar}, Management and Staff profiles diverge clearly on Formality, Warmth, and Questioning, confirming meaningful inter-role variation. Finally, Style accounts for only 7.3\% of all QA items (176/2{,}400), and our $2{\times}2$ fact-vs-style adversarial distractor design prevents systems from exploiting distributional priors.
  \subsection{Dialogue Distribution}

  \begin{table}[t]
  \centering
  \resizebox{\linewidth}{!}{%
  \begin{tabular}{lrrrr}
  \toprule
  \textbf{Topic} & \textbf{Participants} & \textbf{Messages} & \textbf{Sub-tasks} & \textbf{QA} \\
  \midrule
  Technology          & 39 & 10,222 & 436 & 488 \\
  Operations          & 36 & 10,106 & 408 & 471 \\
  Marketing           & 45 & 10,638 & 429 & 467 \\
  Financial Services  & 32 & 10,003 & 423 & 481 \\
  Governance          & 46 & 10,054 & 424 & 493 \\
  \midrule
  \textbf{Total}      & \textbf{170} & \textbf{51,023} & \textbf{2,120} & \textbf{2,400} \\
  \bottomrule
  \end{tabular}%
  }
  \caption{Dialogue statistics across five project topics. \textit{Sub-tasks} denote the number of distinct work tasks respectively.}
  \label{tab:dialogue_stats}
  \end{table}

  The dialogue corpus spans \textbf{5 topics}, each simulating approximately one year (Jan--Dec 2025) of enterprise group-chat communication. As
  summarized in Table~\ref{tab:dialogue_stats}, the dataset contains \textbf{51,023 messages} comprising over \textbf{2.1 million words}, with a
  consistent average message length of 41.5 words across all topics.

  \paragraph{Multi-Theme Structure.}
  Each topic is organized into \textbf{3 concurrent project groups}, with 32--46 participants communicating over 250--255 workdays. Conversations revolve
   around a rich set of \textbf{834--966 distinct themes} per topic (4,556 in total), composed of two sources: (1)~\textit{work tasks} (408--436 per
  topic), representing ongoing project activities such as feature development, code reviews, and design discussions; and (2)~\textit{injected news
  events} (405--543 per topic), simulating real-world information that employees discuss organically---covering industry trends, policy updates, and
  current affairs. This dual-source design ensures that the dialogue corpus captures both structured work-related memory and loosely-structured
  world-knowledge discussions, posing a realistic challenge for memory systems that must distinguish and retrieve from heterogeneous conversational
  threads.

   \paragraph{QA Benchmark.}                                                                                                                       
  Each topic is accompanied by a curated set of question--answer pairs, totaling \textbf{2,400} across all five topics after filtering. QA pairs are                 
  organized into \textbf{three evaluation dimensions} encompassing \textbf{nine sub-tasks}:                                                                          
  \begin{itemize}                                                                                                                                                    
      \item \textbf{Fine-grained Recall} (762 pairs, 31.8\%): Single-hop Retrieval (213), Multi-hop Trajectory (249), and Temporal Duration (300).
      \item \textbf{Memory Awareness} (1,097 pairs, 45.7\%): Constraint (402), Proactivity (427), and Update (268).
      \item \textbf{Profile Understanding} (541 pairs, 22.5\%): Style (176), Skill (169), and Role (196).
  \end{itemize}
 Each QA pair includes an average of 6.7 evidence references pointing to specific messages in the dialogue history, enabling fine-grained evaluation of             
  both retrieval precision and comprehension-synthesis capability.                                                                                                   
  Fine-grained Recall questions adopt an open-ended format because each question targets a well-defined entity (a specific link, date, or person), making the gold   
  answer unambiguous and straightforward to verify.                                                                                                                  
  Memory Awareness and Profile Understanding questions, by contrast, assess generalization and contextual understanding of stored memories---capabilities whose      
  correct answers are more nuanced and harder to judge automatically. To minimize variance introduced by LLM-as-a-Judge scoring, we adopt a multiple-choice format   
  (A/B/C/D) for these two dimensions, where both questions and distractors are carefully crafted to prevent information leakage from the question stem and to ensure
  that every wrong option remains plausible rather than serving as a strawman (see \S\ref{appendix:qa} for distractor design details).

\section{Data Construction Details}\label{appendix:construction}

\subsection{Blueprint Generation}\label{appendix:blueprint}

The blueprint pipeline proceeds through six stages, each followed by automated validation. Below we detail each stage.

\paragraph{Stage 1: Project Topic Generation.}
We begin by generating five project topics spanning diverse enterprise domains---Technology, Operations, Marketing, Financial Services, and Governance---each characterized by a sector-specific context, three to five core challenges, and a set of expected stakeholder roles. Gemini-2.5-Pro first produces a broad pool of candidates; to steer diversity, we provide contrasting examples: desirable topics cross sector boundaries (smart manufacturing, cross-border e-commerce, telemedicine), while near-duplicate clusters (e.g., ``intelligent customer service / intelligent recommendation / intelligent analytics'') are explicitly excluded. Human annotators then screen, merge, and refine the candidates, adjusting scope and resolving overlaps until each surviving topic can sustain a year-long simulated development cycle.

\paragraph{Stage 2: Sub-Project Decomposition.}
Each topic is decomposed into three sub-projects, yielding $5 \times 3 = 15$ distinct units. A Financial Services project, for example, might split into a Risk Assessment Engine, a Transaction Processing System, and a Compliance Reporting Platform. The key constraint is \emph{mutual exclusivity}: sub-projects must be independently executable, so we forbid parent--child overlaps (e.g., ``develop the manufacturing execution system'' alongside ``develop its production scheduling module''). A post-generation validator verifies that no sub-project subsumes another. This structure creates natural cross-group information flow---decisions in one sub-project (say, a risk scoring methodology) constrain design choices in another (transaction validation rules)---while keeping each group chat channel self-contained with its own team roster and task backlog.

\paragraph{Stage 3: Team Member Selection.}
For each sub-project we assemble a team from the shared pool of 170 employees across seven departments (1 Executive, 5 Managers, 164 Staff). The LLM receives the complete employee registry---skill profiles $\mathbf{s}_e$ with proficiency levels and 8-dimensional communication styles $\mathbf{c}_e$---and selects members by balancing skill--task alignment, communication complementarity, and hierarchical coverage. A validator then enforces that every team spans all three ranks and falls within the configured size range. On average, each project involves 37.6 participants.

\paragraph{Stage 4: Communication Style Adaptation.}
Because communication behavior shifts with team context, we adjust each member's style profile after assembly. Given the team's rank distribution and size, the LLM shifts individual dimensions by up to two levels and records the rationale; original profiles are preserved.

\paragraph{Stage 5: Subtask Generation and Sequencing.}
For each sub-project we generate a set of subtasks $\mathcal{T}_{p,j}$ organized into six sequential development phases: Strategy \& Planning, Requirements \& Design, System Architecture \& Tech Selection, Development \& Integration, Testing \& QA, and Deployment \& Operations.

The main challenge here is granularity. A coarse task like ``conduct requirements research'' is insufficient; it must be broken into concrete units such as ``survey target user personas,'' ``collect competitor feature lists,'' and ``interview internal business departments.'' Likewise, ``design the database'' becomes per-table schema design, indexing strategy, and documentation. We supply extensive positive and negative examples to calibrate the LLM's decomposition depth. Each resulting subtask specifies a concrete deliverable, one to three required skills drawn from the skill universe, a development phase, and a dependency type. Tasks are topologically sorted so that architectural decisions precede implementation, API contracts precede integration, and testing follows feature completion. Deadlines are spread across the full calendar year, with early phases concentrated in Q1--Q2 and later phases in Q3--Q4.

\paragraph{Stage 6: Subtask-to-Member Assignment.}
Each subtask is assigned to exactly one team member. Given the ordered subtask list and each member's skill profile, the LLM matches tasks to people by considering proficiency alignment (strong $>$ medium $>$ low), workload balance, and role appropriateness (Executive handles strategy; Managers oversee architecture and coordination; Staff execute implementation and testing). Every assignment comes with a short natural-language rationale that serves as an audit trail. The validator ensures that every member receives at least five subtasks---guaranteeing meaningful dialogue participation---that no subtask is left unassigned, and that temporal feasibility constraints are met.

\paragraph{Cross-Group Dependency Injection.}
After individual blueprints are complete, we inject cross-project dependencies to create the multi-group structure central to EverMemBench. Within each project the three sub-projects already share personnel (especially Managers and the Executive) and technical decisions, producing natural coupling. We formalize three types of dependency: \emph{data contracts}, where one sub-project defines a schema that another consumes and must therefore wait for; \emph{shared infrastructure}, where multiple sub-projects rely on a common component (e.g., an authentication service) managed by one team; and \emph{policy decisions}, where a technical standard adopted in one sub-project constrains choices elsewhere (e.g., the state management library or deployment platform). Each dependency is recorded as a tuple $(P_i, P_j, \text{type}, \text{artifact})$, enabling downstream dialogue generation to reference cross-group decisions and supporting questions that require multi-group evidence chaining.

\paragraph{Blueprint Validation.}
Beyond the per-stage validators described above, a final global check ensures end-to-end consistency before dialogue generation begins. We verify five invariants: (1)~\emph{completeness}---every subtask has an assigned owner, a deadline, and required skills; (2)~\emph{temporal consistency}---no task depends on one scheduled later; (3)~\emph{skill coverage}---every required skill appears in at least one team member's profile at medium proficiency or above; (4)~\emph{load balance}---no employee carries an unreasonable number of concurrent tasks in any 30-day window; and (5)~\emph{role coherence}---strategic tasks are assigned to Executives or Managers, not junior staff. Any failure triggers targeted regeneration of the offending stage.

The resulting blueprints serve as the executable specification for dialogue generation: every conversation references specific subtasks, respects the defined dependencies, and exhibits role-appropriate communication patterns grounded in the blueprint's organizational structure.

\subsection{Conversation Generation}\label{appendix:conversation}

Conversation generation transforms the static blueprints into dynamic, day-by-day multi-party dialogues spanning a full simulated year ($D=365$ days). As introduced in the main text, each sub-project maintains its own group chat channel, and each day's dialogue is generated in a single forward pass conditioned on hierarchical summaries and persona profiles. Dialogues are generated only on workdays; weekends contain no conversations, mirroring the schedule of a real enterprise. This appendix expands on the task scheduling pipeline that bridges blueprints to daily generation, the three-layer summarization mechanism, and the multi-level quality assurance system.

\paragraph{Task Schedule Preparation.}
Before dialogue generation begins, the blueprint's subtask assignments must be converted into a concrete daily schedule that determines which tasks are active on each simulated day. This proceeds in three steps. First, we extract all subtasks from the completed blueprints, adjust deadlines that fall on weekends to the nearest following workday, and deduplicate tasks that appear in overlapping sub-project assignments. Second, Gemini-2.5-Pro~\citep{comanici2025gemini25pushingfrontier} estimates the working-day duration of each task, taking into account its complexity, phase, and inter-task dependencies. Third, a backward-scheduling algorithm computes each task's start date by subtracting the estimated duration from its deadline, automatically resolving conflicts when a predecessor's deadline falls after a dependent task's computed start date. The resulting schedule classifies every task on each workday into one of three categories: \emph{starting} (first day of work), \emph{ongoing} (in progress but not yet due), and \emph{ending} (deadline reached). This three-way partition drives the dialogue generator's expectations about what each day's conversation should contain (new task kickoffs, progress updates, or completion announcements), and is enforced by downstream validation.

\paragraph{Daily Dialogue Synthesis.}
For each workday $d$, the dialogue generator receives five categories of input: (i)~the day's task schedule with its starting/ongoing/ending classification and the identities of each task's assignees, (ii)~the most recent weekly summary $W_{p,j}^{(d)}$, (iii)~the current monthly summary $M_{p,j}^{(d)}$, (iv)~outstanding leader instructions $L_{p,j}^{(d)}$ from the past 30 days, and (v)~the full persona profiles $\{\pi_e\}_{e\in\mathcal{E}_{p,j}}$ of all group members, including their team-adapted communication styles from Stage~4 of the blueprint pipeline.

The generator produces a structured multi-party dialogue where each utterance is attributed to a specific speaker, timestamped within work hours (09:00--18:00), and bound to the task identifiers it discusses. Two constraints are particularly important. First, every task assignee must participate in the discussion of their assigned tasks, since a passive team member who never speaks about their own work would be unrealistic and would deprive downstream QA generation of evidence material. Second, strict deadline discipline is enforced: only tasks whose deadline falls on the current day may be marked as completed, while ongoing tasks may report progress but not closure. This constraint is essential for creating the temporal revision patterns that our evaluation probes; without it, premature completion declarations would collapse the version-tracking complexity needed for temporal reasoning questions.

\paragraph{Hierarchical Summarization.}
Generating coherent dialogue over a full calendar year under LLM context limits requires compressing history without losing the structural cues that support temporal reasoning. As formalized in the main text, we employ three complementary summarization layers.

The \emph{weekly summary} $W_{p,j}^{(d)}$ is generated at the end of each week from the preceding days' dialogues. It organizes progress by task, recording dated milestones in chronological order and noting planned next steps. This provides the dialogue generator with fine-grained recent context (what was decided earlier in the week, what remains blocked, what is expected next), enabling natural continuity across week boundaries.

The \emph{monthly summary} $M_{p,j}^{(d)}$ is generated on the last workday of each month from all dialogues and task records within that period. It groups information by project, listing completed tasks with their completion dates, in-progress items with current status, key milestones, and accumulated factual details such as budgets, team sizes, and technical decisions. Where the weekly summary captures operational momentum, the monthly summary provides the long-range backdrop that prevents the dialogue from contradicting decisions made weeks earlier.

The \emph{leader instruction log} $L_{p,j}^{(d)}$ extracts and retains directives from senior personnel (Executive and Managers) over a rolling 30-day window. Each instruction is categorized as a strategic decision, an operational directive, or a methodological guidance, and is linked to the affected tasks. This mechanism ensures that downstream conversations comply with, not merely mention, the decisions of leadership, creating the policy-adherence patterns targeted by our Memory Awareness evaluation dimension.

\paragraph{Multi-Level Quality Assurance.}
Each day's generated dialogue passes through four sequential validation levels before being accepted, extending the produce--verify philosophy introduced in the blueprint pipeline to the more complex domain of natural-language dialogue.

\emph{Level~1} applies deterministic programmatic rules: timestamps must fall within work hours, speakers must be authorized members of the group they appear in, group names must match the project structure exactly, and raw task identifiers must not leak into dialogue text (task references are maintained through structured metadata bindings). These checks are instantaneous and catch mechanical formatting errors before more expensive semantic analysis begins.

\emph{Level~2} uses an LLM to verify task-completion semantics against the day's schedule. Starting tasks must show evidence of initiation; ending tasks must show evidence of completion; and critically, no task may exhibit premature completion. The checker draws a semantic distinction between forward-looking statements (``I will finish this today,'' ``submitting version 1.0 for review'') and genuine closure markers (``this task is now complete, deliverables archived''), since conflating the two would undermine the temporal integrity of the data.

\emph{Level~3} evaluates logical coherence: an LLM examines the conversation for internal self-contradictions, inconsistencies with information established in weekly and monthly summaries, and violations of recorded leader directives. This check operates with a deliberately lenient threshold, flagging only substantive errors that would make QA items unanswerable or ambiguous.

\emph{Level~4} verifies communication quality against persona profiles: appropriate forms of address (subordinates using formal titles for superiors), absence of self-referencing errors (speakers unnaturally mentioning their own names), alignment with each speaker's 8-dimensional communication style, and role-appropriate expertise in technical discussions.

\paragraph{Bounded Regeneration.}
When validation fails at any level, the system attempts targeted repair rather than regenerating the entire day's dialogue from scratch. A dedicated fix agent receives the flagged dialogue along with a structured conflict report detailing each violation by category and severity. The agent applies minimal, conservative edits, preferring deletion over modification to avoid introducing new errors, such as removing utterances from unauthorized speakers, adjusting out-of-range timestamps, converting premature completion claims into progress statements, and correcting task-identifier bindings. The repaired dialogue is then re-checked through the full four-level pipeline. If validation still fails after two repair attempts, the date is flagged for manual review. All repairs are logged with before-and-after diffs to maintain a complete audit trail and support post-hoc analysis of common generation failure modes.

\paragraph{Cross-Group Coherence.}
Within each project, the three sub-project group chats share personnel, particularly Managers and the Executive, who participate in multiple channels simultaneously. The generation pipeline maintains coherence across groups by ensuring that shared members cannot be speaking in two groups at the exact same timestamp, and that technical decisions referenced across groups remain consistent. When the dialogue generator encounters a task that depends on a cross-group artifact (a data contract, shared infrastructure component, or policy decision injected during the blueprint phase), it conditions the conversation on the relevant cross-group context so that references to external decisions are accurate and temporally appropriate. This mechanism is what makes multi-hop, cross-group QA items possible: the evidence trail genuinely spans multiple group chats rather than being artificially duplicated.

\subsection{Q\&A Generation}\label{appendix:qa}

We synthesize 2,400 QAE triples $(q,a,e)$ through three specialized pipelines, one per evaluation dimension. All three pipelines share a common principle we call the \emph{missing-key rule}: each question provides a scenario (the ``lock'') whose resolution depends on specific evidence buried in the dialogue history (the ``key''), and the question itself must never leak the key. This principle guides both question design and distractor construction across all dimensions. Below we detail the generation strategy for each dimension.

\paragraph{Fine-grained Recall.}
This dimension evaluates precise retrieval of entities, timestamps, and events from dense multi-turn discussions. We organize generation into sub-branches of increasing difficulty. \emph{Single-hop retrieval} targets direct entity grounding: extracting a specific deliverable link, a stated deadline, or a named assignee from a particular discussion thread. \emph{Duration and interval} questions require computing time spans, either the working-day length of a single task or the gap between two events (e.g., from one task's completion to the next task's initiation by the same person). \emph{Existence checks} include both positive cases (verifying that a stated event did occur) and negative cases designed to test hallucination resistance, where we construct near-miss scenarios referencing plausible but non-existent events. \emph{Multi-hop retrieval} requires chaining evidence across speakers or groups: identifying a person through a compositional constraint (e.g., ``the colleague responsible for writing the Swagger documentation for authentication''), then tracing that person's timeline to a subsequent event. When the base narrative lacks the specific evidence structure needed for a multi-hop chain, we employ controlled implantation via \emph{field-bridge decoupling}: a target reasoning chain $R_o$ is decomposed into a bridge segment $R_2$ that maps a scenario to an intermediate identifier and a dictionary segment $R_3$ that resolves the identifier to a concrete solution. These segments are embedded into the dialogue history as plausible meeting notes or data-dictionary entries authored by real project members; seeing $R_3$ alone reveals no connection to the original scenario, so the solver must retrieve $R_2$ to complete the chain. To create a realistic difficulty gradient, we additionally extract a decoy rule $R_1$ from existing dialogue, representing a general default procedure that the question naturally evokes. The question references $R_1$'s context but includes a subtle special condition that triggers the override path through $R_2$ and $R_3$, forcing the system to recognize that the general rule does not apply and to follow the multi-hop chain instead.

For all sub-branches, we apply \emph{structure mining}: an LLM agent traverses the blueprint and dialogue structure to identify naturally occurring retrieval targets, such as task completion markers, cross-group decision references, and personnel role bindings. To ensure diversity, we use embedding-based similarity filtering to prevent semantically redundant questions from clustering around the same dialogue segments.

\paragraph{Memory Awareness.}
This dimension tests whether the system can mobilize past information to solve new problems, not merely recall it. Generation follows a five-step pipeline for each QA item. In the first step, the generator selects one to four dialogue snippets as evidence, focusing on segments that contain logical anchors: constraint definitions, contradiction points, or high-stakes directives. In the second step, a verification agent checks factual accuracy, logical soundness, and evidence completeness. In the third step, the generator produces the question, the correct answer, and three distractors. In the fourth step, a chain-of-thought auditor checks whether the question is answerable without evidence (which would indicate information leakage), whether all options appear equally authoritative, and whether length imbalance reveals the correct answer. In the fifth step, difficulty is upgraded through adversarial perturbations: keyword substitution replaces explicit technical terms with functional descriptions to force semantic rather than lexical retrieval; parameter removal strips explicit constraint mentions from the question so that the solver must infer them from evidence; and honey-trap options are added that satisfy common sense but violate specific rules established in the dialogue.

The three sub-tasks target distinct cognitive operations. \emph{Constraint} questions present novel scenarios that require applying implicit rules extracted from past discussions (e.g., recognizing that a data-contract owner, not the downstream consumer, must approve schema changes). \emph{Proactivity} questions simulate situations where the user issues an instruction that conflicts with a previously established hard rule; the system must detect the violation and provide a reminder rather than blindly executing. \emph{Updating} questions require resolving chronological precedence: initial specifications are later revised, and the system must identify the current valid state rather than returning stale information.

\paragraph{Profile Understanding.}
This dimension examines whether the system can aggregate distributed signals into stable user models. We employ a $2\times 2$ fact-versus-style matrix to construct distractors that disentangle factual correctness from persona alignment. For each question, we generate four options: the correct answer (fact correct, style correct), F1 (fact correct, style wrong), F2 (fact wrong, style correct), and FF (fact wrong, style wrong). The critical distractor is F2, which perfectly mimics the target person's communication style while containing fabricated factual content. F2 is deliberately made 10--20\% longer than the correct answer to exploit the tendency of language models to prefer more detailed options. FF reuses F2's sentence structure with only the style expressions swapped, preventing it from being dismissible as a strawman.

Factual errors in F2 and FF follow five misleading strategies: quoting real names and timestamps from the evidence but fabricating conclusions (half-true-half-false); attributing one person's work to another (role misattribution); reversing temporal states such as ``completed'' versus ``in progress'' (temporal reversal); inverting cause-effect relationships (causal inversion); and subtly modifying quantities or degrees (degree manipulation). All fabricated facts must remain professionally plausible, representing alternative reasonable decisions rather than obvious technical absurdities.

The three sub-tasks probe different facets of user modeling. \emph{Style} questions ask the system to draft a reply on behalf of a specific person, where the correct option matches both the factual record and the person's characteristic communication register (formality, emoji usage, humor). \emph{Skill} questions require recommending a technical solution that aligns with both the speaker's personal expertise and the team's ratified technology stack. \emph{Role} questions test whether improvement suggestions are framed from the speaker's professional perspective (e.g., a developer proposing pipeline reviews rather than test-case documentation).

\subsection{Q\&A Quality Control}\label{appendix:quality}

Quality control enforces three properties: non-triviality (items cannot be solved without context), soundness (items can be solved with the designated evidence), and uniqueness (items cannot be solved from unintended evidence segments). We implement these through a three-phase filter that deliberately favors conservative retention.

\subsubsection{Blind Test}\label{appendix:blind}

The blind test eliminates items solvable through parametric knowledge or annotation artifacts. Each QA item is stripped of all dialogue context and evidence, leaving only the question and answer options, and submitted in parallel to four frontier LLMs (Gemini-2.5-Pro, Claude Sonnet 4.5, GPT-5.1, and Qwen3-235B). Each model returns a predicted answer, a confidence score, and a reasoning chain.

We classify results into four scenarios. \emph{Scenario A} (serious leakage): a majority of models answer correctly with high confidence, indicating that the question is solvable from world knowledge alone; these items are rejected. \emph{Scenario B} (random guessing): answers are dispersed with low confidence, confirming that context is required; these items are retained. \emph{Scenario C} (strong distractors): a majority of models answer incorrectly, suggesting that the distractors are effective; these are marked as high-quality items. \emph{Scenario D} (premium difficulty): only one or two models answer correctly with sophisticated reasoning, indicating that the question is challenging but fair. Only Scenario A items are automatically rejected; all others proceed to Phase II.

\subsubsection{Evidence Grounding}\label{appendix:grounding}

Evidence grounding verifies that each item is both answerable from its designated evidence and not answerable from any other dialogue segment. We partition the full dialogue corpus $\mathcal{C}_p$ into segments aligned with subtask boundaries from the blueprint. For each QA item, the segment containing its evidence is designated as $S^+$ (positive context) and all remaining segments as $S^-$ (negative contexts).

The \emph{sufficiency test} provides an LLM with the question, options, and $S^+$, requiring it to derive the correct answer through explicit reasoning. The model must extract specific dialogue citations (speaker, timestamp, content) supporting its choice. If the model fails to answer correctly or selects ``cannot determine,'' the item is flagged as insufficiently grounded. The test also records which portions of the original evidence the model actually used, providing an R-coverage metric that measures how much of the annotated evidence is necessary versus redundant.

The \emph{uniqueness test} checks whether any negative segment accidentally supports the correct answer. For each $S^- \in \mathbf{S} \setminus \{S^+\}$, we conduct a two-phase probe. In the first phase (fast screening), the model is forced to select among A/B/C/D without a ``cannot determine'' option; if it selects incorrectly, the segment is cleared. If it selects correctly, a second phase provides the ``cannot determine'' option and requests detailed reasoning. Items where the model still selects the correct answer in the second phase are flagged as having uniqueness issues, since the evidence for that answer exists in an unintended location.

Items that fail sufficiency are rejected outright. Items with uniqueness warnings are escalated to human review, where annotators determine whether the unintended evidence path represents a genuine alternative or a false positive from the LLM's reasoning.

\subsubsection{Human Audit}\label{appendix:human}

The final phase addresses residual logical and pragmatic issues that automated filters cannot reliably detect. Items are prioritized by risk level: those flagged by Phase II receive priority review, items that passed both automated phases receive spot checks. Human reviewers evaluate four dimensions: question reasonability (does the question follow natural human inquiry patterns?), answer rigor (is the correct answer unambiguous and fully evidence-based?), distractor deceptiveness (are wrong options genuinely challenging rather than obviously implausible?), and evidence sufficiency (does the annotated evidence fully support the reasoning chain?). Items that fail any dimension are either revised and re-tested through the full pipeline or rejected entirely. This conservative approach ensures that benchmark failures can be confidently attributed to system limitations rather than annotation noise.

\subsubsection{Post-Hoc Validation}\label{appendix:zeroshot}

As a final check, we verify that the curated benchmark remains non-trivial after all filtering stages. We evaluate GPT-4.1-Mini in a \emph{zero-context} setting: the model receives only the question and answer options with no dialogue history or retrieved memories. As shown in Table~\ref{tab:zero_shot}, performance across all nine sub-tasks remains at or below chance level (25\% for four-way multiple choice), with Multi-Hop retrieval as low as 2.01\%. This end-to-end result complements the construction-time blind test (\S\ref{appendix:blind}): whereas the blind test filters out items answerable by frontier LLMs \emph{before} inclusion, the zero-context baseline confirms that the final item set cannot be solved without access to the underlying conversation corpus.

\begin{table}[H]
\centering
\small
\setlength{\tabcolsep}{3pt}
\renewcommand{\arraystretch}{1.1}
\resizebox{\linewidth}{!}{%
\begin{tabular}{l *{3}{c} *{3}{c} *{3}{c}}
\toprule
\multirow{2}{*}{\textbf{Model}}
& \multicolumn{3}{c}{\textbf{Fine-Grained Recall}}
& \multicolumn{3}{c}{\textbf{Memory Awareness}}
& \multicolumn{3}{c}{\textbf{Profile Understanding}} \\
\cmidrule(lr){2-4}\cmidrule(lr){5-7}\cmidrule(lr){8-10}
&
\textit{Single} &
\textit{Multi} &
\textit{Temp.} &
\textit{Const.} &
\textit{Proac.} &
\textit{Upd.} &
\textit{Style} &
\textit{Skill} &
\textit{Role} \\
\midrule
GPT-4.1-Mini & 15.02 & 2.01 & 12.67 & 11.69 & 10.07 & 7.84 & 17.61 & 26.04 & 26.53 \\
\bottomrule
\end{tabular}%
}
\caption{Zero-context performance of GPT-4.1-Mini without any memory access. All scores remain at or below chance level, confirming that questions require genuine memory retrieval.}

\label{tab:zero_shot}
 \vspace{-20pt}
\end{table}

\section{Evaluation Details}\label{appendix:evaluation}

\subsection{Answer Prompts}\label{appendix:answer}

We use separate prompts for multiple-choice and open-ended questions. For memory-augmented systems, retrieved memories serve as context (Figures~\ref{fig:memory_mc_prompt}--\ref{fig:memory_oe_prompt}); for the full-context LLM baseline, the complete dialogue transcript is provided directly (Figures~\ref{fig:fullcontext_mc_prompt}--\ref{fig:fullcontext_oe_prompt}).

\subsection{LLM-as-a-Judge Prompt}\label{appendix:judge}

We adopt an LLM-as-a-Judge protocol for open-ended answer evaluation. The judge receives the question, gold answer, and generated answer, and returns a binary CORRECT/WRONG label with a one-sentence rationale (Figure~\ref{fig:prompt_judge}).

\subsection{LLM-as-Judge Reliability }\label{sec:appendix_b}

We randomly selected 30 non-overlapping open Q\&A pairs from EverMemBench, and generated model answers for each question. We recruited annotators via Prolific. For each Q\&A pair, five independent human evaluators judged whether the generated answer was correct given the question and the reference answer. All participants provided informed consent via the platform interface and were compensated at approximately \$12.00/hour, consistent with fair-pay guidelines for academic research and above local minimum wage standards. Table~\ref{tab:alignment} shows strong agreement between the LLM-as-judge protocol and human annotations. These results suggest that LLM-as-Judge achieves human-level reliability for answer verification, enabling evaluation that is rigorous, reproducible, and cost-efficient.

\begin{table}[H]
\centering
\small
\setlength{\tabcolsep}{4pt}
\begin{tabular}{lcccc}
\toprule
System & Cohen's $\kappa$ & 95\% CI & Accuracy & Pearson $r$ \\
\midrule
MemOS     & 0.927 & [0.756, 1.000] & 96.7\%  & 0.929 \\
MemoBase  & 0.930 & [0.756, 1.000] & 96.7\%  & 0.932 \\
\bottomrule
\end{tabular}
\caption{Reliability matrix for LLM-as-Judge.}
\label{tab:alignment}
 \vspace{-25pt}
\end{table}

\section{Additional Experimental Results}\label{appendix:exp}
\subsection{Performance across Different Topics}\label{appendix:topics}

\begin{table}[htbp]
\centering
\small
\setlength{\tabcolsep}{5pt}
\renewcommand{\arraystretch}{1.1}

\begin{tabular}{l ccccc}
\toprule
\textbf{System } & \textbf{T1} & \textbf{T2} & \textbf{T3} & \textbf{T4} & \textbf{T5} \\
\midrule
LLM & 38.98 & 35.43 & 37.79 & 35.98 & 38.90 \\
Zep & 40.00 & 39.15 & 36.37 & \textbf{43.93} & 40.50 \\
Mem0 & 35.97 & 30.02 & \textbf{42.49} & 37.23 & 39.99 \\
MemOS & \textbf{40.73} & \textbf{41.05} & 42.03 & 41.58 & \textbf{43.21} \\
MemoBase & 32.51 & 28.57 & 31.25 & 30.83 & 33.26 \\
\bottomrule
\end{tabular}

\caption{Performance across five project domains (T1--T5: Technology, Operations, Marketing, Financial Services, Governance). Bold indicates the best system for each topic.}
\label{topics}
 \vspace{-25pt}
\end{table}

Table~\ref{topics} reports performance across five project domains. System rankings are largely stable: MemOS leads on three of five topics (T1, T2, T5) and places second on the remaining two, while MemoBase ranks last on all five domains, indicating that capability gaps are fundamental rather than domain-dependent. Cross-domain average accuracy ranges from 34.8\% (T2) to 39.2\% (T5), confirming that no single domain is systematically easier or harder. Cross-domain stability varies notably across systems: MemOS shows the smallest performance spread ($\Delta$=2.5pp), while Mem0 exhibits the largest ($\Delta$=12.5pp, dropping from 42.49\% on T3 to 30.02\% on T2), suggesting that some retrieval strategies are more sensitive to domain-specific information structures than others.
 \begin{figure*}[htbp]
\centering
\begin{tcolorbox}[
    colback=gray!5,
    colframe=black!70,
    title={\textbf{Memory-Augmented Answer Prompt --- Open-Ended}},
    fonttitle=\bfseries\small,
    fontupper=\footnotesize,
    boxrule=0.8pt,
    arc=2pt]
You are an intelligent memory assistant tasked with retrieving accurate information from conversation memories. You will be given retrieved memories from a multi-person group chat and one open-ended question.
Your task is to answer the question using ONLY the information in the memories.

\# INSTRUCTIONS:\\
1. Carefully analyze all provided memories from the group chat.\\
2. Pay special attention to timestamps to determine when events occurred.\\
3. If the question asks about a specific event or fact, look for direct evidence in the memories.\\
4. If memories contain contradictory information, prioritize the most recent memory.\\
5. If the question involves time references (like ``last year'', ``two months ago''), calculate the actual date based on the memory's timestamp.\\
6. Always convert relative time references to specific dates, months, or years in your answer.\\
7. Pay attention to who said what --- the memories may involve multiple participants.\\
8. The answer should be concise and specific (under 5--6 words when possible).\\
9. Do NOT output any reasoning steps, explanations, or extra text beyond the final answer.

\# APPROACH (Think step by step internally):\\
1. Examine all memories that contain information related to the question.\\
2. Examine timestamps and content carefully.\\
3. Look for explicit mentions of dates, times, locations, or events that answer the question.\\
4. If the answer requires calculation (e.g., converting relative time references), do so.\\
5. Formulate a precise, concise answer based solely on the evidence in the memories.

Output format (must be followed exactly):\\
Output ONLY the answer text

[MEMORIES]
\{context\}

[QUESTION]
\{question\}
\end{tcolorbox}
\vspace{-1em}
\caption{Memory-augmented answer prompt for open-ended questions.}
\label{fig:memory_oe_prompt}
\end{figure*}

 \begin{figure*}[htbp]
\centering
\begin{tcolorbox}[
    colback=gray!5,
    colframe=black!70,
    title={\textbf{Memory-Augmented Answer Prompt --- Multiple-Choice}},
    fonttitle=\bfseries\small,
    fontupper=\footnotesize,
    boxrule=0.8pt,
    arc=2pt]
You are a rigorous question-answering assistant. You will be given retrieved memories from
a multi-person group chat and one multiple-choice question with four options.
Your task is to choose the single best answer (A/B/C/D) based ONLY on the provided memories.

Rules:\\
1. Use only information explicitly stated in the memories or directly entailed by them.
Do NOT use outside knowledge, assumptions, or guesses beyond the memories.\\
2. The memories come from group chat conversations involving multiple participants.
Pay attention to who said what and when.\\
3. If multiple options seem plausible, choose the one most strongly and directly supported
by the memories.\\
4. If the memories do not provide enough information to be certain, you MUST still pick one
option (A/B/C/D). Choose the option that is least inconsistent with the memories.\\
5. Pay special attention to timestamps to determine when events occurred.\\
6. If memories contain contradictory information, prioritize the most recent memory.\\
7. Do NOT output any reasoning, explanation, punctuation, or extra text.

Output format (must be followed exactly):\\
Output ONLY a single uppercase letter: A or B or C or D

[MEMORIES]
\{context\}

[QUESTION]
\{question\}

[OPTIONS]
\{options\}
\end{tcolorbox}
\vspace{-1em}
\caption{Memory-augmented answer prompt for multiple-choice questions.}
\label{fig:memory_mc_prompt}
\end{figure*}

 \begin{figure*}[htbp]
\centering
\begin{tcolorbox}[
    colback=gray!5,
    colframe=black!70,
    title={\textbf{Full-Context Answer Prompt --- Open-Ended}},
    fonttitle=\bfseries\small,
    fontupper=\footnotesize,
    boxrule=0.8pt,
    arc=2pt]
You are a rigorous question-answering assistant. You will be given a very long dialogue transcript and one open-ended question. Your task is to answer the question using ONLY the information in the dialogue.

Rules:\\
1. Use only information explicitly stated in the dialogue or directly entailed by it. Do NOT use outside knowledge, assumptions, or speculation.\\
2. If the dialogue does not contain enough information to answer, output exactly: NOT ENOUGH INFORMATION\\
3. Keep the answer concise, specific, and aligned with the dialogue. Do not add commentary.\\
4. Do NOT output any reasoning steps, explanations, or extra text beyond the final answer.

Output format (must be followed exactly):\\
Output ONLY the answer text (or exactly: NOT ENOUGH INFORMATION)

[DIALOGUE]\\
\{context\}

[QUESTION]\\
\{question\}
\end{tcolorbox}
\caption{Full-context answer prompt for open-ended questions. The complete dialogue transcript is provided directly to the LLM without a memory system.}
\label{fig:fullcontext_oe_prompt}
\end{figure*}

 \begin{figure*}[htbp]
\centering

\begin{tcolorbox}[
    colback=gray!5,
    colframe=black!70,
    title={\textbf{Full-Context Answer Prompt --- Multiple-Choice}},
    fonttitle=\bfseries\small,
    fontupper=\footnotesize,
    boxrule=0.8pt,
    arc=2pt]
You are a rigorous question-answering assistant. You will be given a very long dialogue transcript and one multiple-choice question with four options. Your task is to choose the single best answer (A/B/C/D) based ONLY on the dialogue.

Rules:\\
1. Use only information explicitly stated in the dialogue or directly entailed by it. Do NOT use outside knowledge, assumptions, or guesses beyond the dialogue.\\
2. If multiple options seem plausible, choose the one most strongly and directly supported by the dialogue.\\
3. If the dialogue does not provide enough information to be certain, you MUST still pick one option (A/B/C/D). Choose the option that is least inconsistent with the dialogue.\\
4. Do NOT output any reasoning, explanation, punctuation, or extra text.

Output format (must be followed exactly):\\
Output ONLY a single uppercase letter: A or B or C or D

[DIALOGUE]\\
\{context\}

[QUESTION]\\
\{question\}

[OPTIONS]\\
\{options\}
\end{tcolorbox}
\vspace{-5pt}
\caption{Full-context answer prompt for multiple-choice questions. The complete dialogue transcript is provided directly to the LLM without a memory system.}
\label{fig:fullcontext_mc_prompt}
\end{figure*}

\begin{figure*}[htbp]
\centering
\begin{tcolorbox}[
    colback=gray!5,
    colframe=black!70,
    title={\textbf{LLM-as-a-Judge --- System Prompt}},
    fonttitle=\bfseries\small,
    fontupper=\footnotesize,
    boxrule=0.8pt,
    arc=2pt]
You are an expert grader that determines if answers to questions match a gold standard answer.
\end{tcolorbox}

\vspace{0.3cm}

\begin{tcolorbox}[
    colback=gray!5,
    colframe=black!70,
    title={\textbf{LLM-as-a-Judge --- User Prompt}},
    fonttitle=\bfseries\small,
    fontupper=\footnotesize,
    boxrule=0.8pt,
    arc=2pt]
Your task is to label an answer to a question as `CORRECT' or `WRONG'. You will be given:\\
(1) a question (about a multi-person group chat),\\
(2) a `gold' (ground truth) answer,\\
(3) a generated answer\\
which you will score as CORRECT/WRONG.

The questions are about events, facts, or details mentioned in multi-person group chat conversations.
The gold answer is usually a concise answer that includes the key information.

For example:\\
Question: What project was announced on January 9th?\\
Gold answer: Carbon Emission Accounting Platform

The generated answer might be longer, but you should be generous with your grading ---
as long as it contains the same key information as the gold answer, it should be CORRECT.

For time-related questions, the gold answer will be a specific date/time.
The generated answer might use different formats (e.g., ``May 7th'' vs ``7 May'' vs ``2025-05-07''),
but as long as it refers to the same date/time, it should be CORRECT.

For the specific window of date, a +/- 1 day difference is acceptable due to timezone processing variations.

For multiple choice questions where the gold answer is a letter (A/B/C/D),
the generated answer should match exactly to be CORRECT.

Now grade this:\\
Question: \{question\}\\
Gold answer: \{golden\_answer\}\\
Generated answer: \{generated\_answer\}

First, provide a short (one sentence) explanation of your reasoning,
then finish with CORRECT or WRONG.
Do NOT include both CORRECT and WRONG in your response.

Return the label in JSON format with the key ``label'': \\
\{``label'': ``CORRECT''\} or \{``label'': ``WRONG''\}
\end{tcolorbox}
\caption{LLM-as-a-Judge prompt for evaluating generated answers.}
\label{fig:prompt_judge}
\end{figure*}

\subsection{Performance Across Different Groups Numbers}\label{appendix:groups}
\begin{table}[h]
  \centering
   \small
  \begin{tabular}{lrrr}
    \toprule
    QAR Group Count &  1 &  2 &  3 \\
    \midrule
    \multicolumn{4}{l}{\textbf{Answer Model: GPT-4.1-mini}} \\
    \midrule
    llm & 0.4247 & 0.2463 & 0.1818 \\
       zep & 0.4743 & 0.2925 & 0.0909 \\
    mem0 & 0.4694 & 0.2449 & 0.1818 \\
    memos & 0.4791 & 0.3333 & 0.2727 \\
      memobase & 0.3702 & 0.2476 & 0.1818 \\
    \midrule
    \multicolumn{4}{l}{\textbf{Answer Model: Llama4-Scout}} \\
    \midrule
    llm & 0.4779 & 0.2585 & 0.0909 \\
     zep & 0.4646 & 0.3034 & 0.0909 \\
    mem0 & 0.4743 & 0.3088 & 0.0909 \\
    memos & 0.4822 & 0.3388 & 0.1818 \\
    memobase & 0.3902 & 0.2707 & 0.0909 \\
       \midrule
    \multicolumn{4}{l}{\textbf{Answer Model: Gemini-3-Flash}} \\
    \midrule
    llm & 0.8627 & 0.5401 & 0.4545 \\
    zep & 0.6576 & 0.4163 & 0.1818 \\
    mem0 & 0.6340 & 0.3850 & 0.1818 \\
    memos & 0.6697 & 0.4177 & 0.1818 \\
    memobase & 0.6564 & 0.3823 & 0.1818 \\
    \bottomrule
  \end{tabular}
  \caption{Accuracy by Grouping across Answer Models}
  \label{acc_combined}
   \vspace{-13pt}
\end{table}
Table~\ref{acc_combined} further breaks down accuracy by the number of cross-group hops required, under three answer models. The cross-group degradation pattern observed is consistent across all answer models: 
 accuracy drops sharply from single-group to three-group questions regardless of backbone capability.


\section{Case Examples}\label{appendix:examples}

\newcommand{\emoji}[1]{\texttt{[:#1:]}}
\captionsetup[figure]{skip=4pt}

In this section, we present representative case examples for each category and provide qualitative analyses based on the typical errors we observed when inspecting the evaluation outputs of existing memory systems. These analyses further highlight the remaining gaps and opportunities for improvement in current memory systems. Figures~\ref{fig:recall_cases_part1}--\ref{fig:profile_un4} report each case in a unified format, including the full question, the correct answer, a typical erroneous answer, and the corresponding evidence. Each sub-task description and the accompanying qualitative error analysis are provided alongside the corresponding case below.

\begin{figure*}[t!]
    \centering
    \begin{casebox}{Single-hop Retrieval}
      \textbf{\mytriangle\ Question:}\\
      In the ``Online Medical Consultation Service System,'' after Yangmeng
      Peng completes the UI design task\dots\ what is the link to the
      delivery page she provided \dots?

      \vspace{0.1cm}
      \textbf{\mytriangle\ Correct Answer:}\\
      \url{https://sd.confluence.com/.../Doctor+Scheduling+UI+Delivery+20250430}

      \vspace{0.1cm}
      \textbf{\mytriangle\ Typical Wrong Answer:} \\
      \url{https://sd.figma.com/.../doctor-scheduling-ui-v2}

      \vspace{0.1cm}
      \textbf{\mytriangle\ Evidence}
      \texttt{(Full context: 2025-04-24 to 2025-04-30, Group 3)}:

      \textit{\textbf{Gold }--- Confluence delivery link (task completion):}\\
      \texttt{[2025-04-30, Group 3, \#19]} Yangmeng Peng:
      ``\dots the UI design task \dots has been completed\dots Here is the
      \textcolor{blue}{Confluence} delivery page link:
     \url{https://sd.confluence.com/.../Doctor+Scheduling+UI+Delivery+20250430}.''

           \textit{\textbf{Distractor:}} A detail-rich Figma link from the same speaker appears 2 days earlier, serving as a strong but incorrect retrieval candidate.

  \end{casebox}
  \vspace{-0.5em}
    \caption{Single-hop Retrieval example in Fine-grained Recall.}
    \label{fig:recall_cases_part1}
  \vspace{-0.5em}
    \centering
    \begin{casebox}{Multi-hop Trajectory}
      \textbf{\mytriangle\ Question:}\\
      In the Carbon Footprint Collaboration System group, how long after
      the colleague responsible for writing the Swagger API documentation
      for the user authentication service completed that task did they
      start their next independent task in this project group?

      \vspace{0.1cm}
      \textbf{\mytriangle\ Correct Answer:}\\
      From May 15, 2025 to August 4, 2025, there is a period of 81 days.

      \vspace{0.1cm}
      \textbf{\mytriangle\ Typical Wrong Answer:} \\
      ``NOT ENOUGH INFORMATION'' or incorrect intervals.

      \vspace{0.1cm}
      \textbf{\mytriangle\ Evidence}
      \texttt{(Full context: 2025-05-09 to 2025-05-15 \& 2025-08-04 to 2025-08-08,
      Group 3)}:

      \textit{Hop 1 --- Person identification
      (API documentation + user authentication
      $\rightarrow$ Jiahui Zhao):}\\
      \texttt{[2025-05-09, Group 3, \#6]} Jiahui Zhao:
      ``\dots today I'm starting the \textcolor{blue}{API documentation}
      for the \textcolor{blue}{user and authentication services}. \dots''

      \textit{Hop 2 --- True task completion anchor
      (End point: May 15):}\\
      \texttt{[2025-05-15, Group 3, \#13]} Jiahui Zhao:
      ``
    \textcolor{blue}{[Task Completed]} The API interface documentation
      for user and authentication services has been updated to the
      \textcolor{blue}{final version}\dots Swagger link:
      https://sd.swagger.io/\dots''

      \textit{Hop 3 --- Next independent task anchor
      (Start point: Aug 4, 81 days later):}\\
      \texttt{[2025-08-04, Group 3, \#3]} \textcolor{blue}{Jiahui Zhao}:
      ``\dots I'm about to sync my plan\dots My preliminary design
      is that when the user clicks the `Submit and Calculate'
      button\dots sending a message\dots to RabbitMQ\dots''
  \end{casebox}
    \vspace{-0.5em}
    \caption{Multi-hop Trajectory example in Fine-grained Recall.}
    \label{fig:recall_cases_part2}
\end{figure*}

\subsection{Fine-grained Recall Examples}\label{appendix:example-recall}

As defined in \S3.1, Fine-grained Recall evaluates whether a system can retrieve precise facts from dense, multi-party discussions where relevant evidence is scattered across speakers and groups. In this subsection, we present one representative case per sub-task and analyze the dominant failure patterns observed across all tested memory systems.
\paragraph{Single-hop Retrieval.}
This case requires the system to find a specific Confluence delivery link provided by a particular team member upon task completion.
The system must accurately locate and extract this precise URL from the conversation history while filtering out
similar but irrelevant links (e.g., intermediate Figma design links). This evaluates \textit{Single-hop Retrieval},
targeting fundamental entity grounding within a specific scope.
As shown in Figure~\ref{fig:recall_cases_part1}, both links share the same speaker (Yangmeng Peng), the same task context (doctor scheduling UI),
and appear only 2 days apart within the same group. All tested memory systems and baseline LLMs
returned the Figma link instead of the Confluence link, revealing three compounding failure factors.
(1)~\emph{Message-length bias}: The Figma message is substantially longer and richer in technical
detail (filtering logic, batch operations), producing a higher embedding similarity to the query;
the Confluence delivery message is brief and administrative, causing it to rank lower despite being
the correct target.
(2)~\emph{Recency-insensitive retrieval}: The Figma link appears two days \textit{before} the
Confluence link. Systems that do not model task lifecycle stages treat both as equally valid
candidates and default to the one with stronger lexical overlap, regardless of temporal ordering.
(3)~\emph{Inability to distinguish artifact types}: Current retrieval pipelines lack a semantic
model of task progression (draft $\rightarrow$ review $\rightarrow$ delivery), and thus cannot
distinguish \textit{intermediate artifacts} (Figma design links) from \textit{final deliverables}
(Confluence delivery pages)---a confusion systematic across all tested systems.

\paragraph{Multi-hop Trajectory.}
This case requires the system to calculate the time interval between a team member's task completion and their next
independent assignment. To answer correctly, the system must identify the person through a compositional constraint,
locate the true completion event while ignoring premature announcements, then trace this individual's timeline across
a prolonged idle period. This evaluates \textit{Multi-hop Trajectory}, requiring the model to reconstruct a specific
individual's work trajectory from information scattered across multiple conversation threads.
As shown in Figure~\ref{fig:recall_cases_part2}, this case layers four distinct retrieval
challenges. \textit{First}, the question never names Jiahui Zhao directly; the system must resolve
the compositional constraint ``Swagger'' $\cap$ ``user authentication'' to identify the correct
person, while other colleagues also discuss Swagger in different contexts. \textit{Second}, an
81-day gap separates the completion event (May~15) from the next task start (Aug~4); the system
must traverse ${\sim}$2.5 months of unrelated group activity to locate the correct anchor.
\textit{Third}, the conversation contains premature completion signals (May~13: ``will complete
today'') that could mislead a system into selecting the wrong end-point. \textit{Fourth}, even if
both dates are correctly retrieved, the model must perform accurate calendar-day arithmetic. Most
memory systems failed at the retrieval stage, returning ``NOT ENOUGH INFORMATION,'' while those
that did retrieve partial evidence sometimes anchored on the wrong completion date.

\paragraph{Temporal Duration.}
This sub-task asks the system to determine the duration of a specific task from initiation to completion.
This evaluates \textit{Temporal Duration}, testing whether the system can correctly extract and compute
a time span from noisy, interleaved conversations. We present two complementary cases in
Figure~\ref{fig:recall_cases_part3}: a \textit{calendar-day} variant and a \textit{working-day} variant.

The calendar-day case (top) reveals two failure factors that separate \textit{retrieval quality} from \textit{reasoning ability}.
(1)~\emph{Keyword collision across speakers}: The phrase ``archived in Confluence'' appears from
multiple team members on nearby dates (Apr~2 by Mingzhi Li, Apr~9 by Huahua Han), creating dense
retrieval noise. Systems relying on keyword or semantic matching over this phrase return these
irrelevant events instead of the correct one (Apr~10 by Luhao Zhao), because the query term
``archiving'' matches all three equally well without person-specific filtering.
(2)~\emph{Endpoint asymmetry}: Most memory systems successfully retrieved the task initiation event
(Apr~4), which carries a distinctive marker (``starting a new task''). However, the archival event
requires jointly matching the correct person, the correct task, \textit{and} a completion signal---a
conjunction that is harder to satisfy than the single-predicate start event. This asymmetry means
systems consistently find one anchor but not the other.

The working-day case (bottom) introduces an additional \textit{arithmetic reasoning} challenge.
Beyond retrieving the correct temporal anchors, the model must recognise that ``working days''
excludes weekends---Apr~22 (Tue) to Apr~28 (Mon) spans 7 calendar days but only 5 working days---a
distinction that demands genuine temporal reasoning rather than simple subtraction. Such
working-day calculations are routine in real-world project coordination, yet they prove
difficult for current systems: most memory systems either conflate calendar and working days
or refuse to answer altogether. Notably, as shown in the oracle analysis (Table ~\ref{tab:oracle}),
even when ground-truth evidence is directly provided, the answer-model accuracy on temporal
working-day questions remains substantially lower than on other question types, confirming that
these questions compound two orthogonal difficulties: (i)~retrieving the correct start/end
anchors and (ii)~performing date arithmetic that accounts for weekends. This makes temporal
duration a meaningful stress test for end-to-end collaborative assistants.

\begin{figure*}[t!]
    \centering
    \begin{casebox}{Temporal Duration --- Calendar Days}
      \textbf{\mytriangle\ Question:}\\
      How many days passed from when Luhao Zhao started working on
      planning the information architecture and sitemap for the
      enterprise energy consumption monitoring system until he completed
      the archiving of this work?

      \vspace{0.1cm}
      \textbf{\mytriangle\ Correct Answer:}\\
      The task started on April 4, 2025, and ended on April 10, 2025,
      lasting 7 days.

      \vspace{0.1cm}
      \textbf{\mytriangle\ Typical Wrong Answer:}\\
      ``264 days,'' ``6 days,'' or ``NOT ENOUGH INFORMATION.''

      \vspace{0.1cm}
      \textbf{\mytriangle\ Evidence}
      \texttt{(Full context: 2025-04-04 to 2025-04-10, Group 2)}:

      \textit{Anchor 1 --- Task initiation (Start point: Apr 4):}\\
      \texttt{[2025-04-04, Group 2, \#3]} Luhao Zhao:
      ``\dots  I'm \textcolor{blue}{starting} a new task \dots: \textcolor{blue}{"Designing the overall information
      architecture and sitemap"} \dots I'll have a
      first draft out this afternoon.''

      \textit{Anchor 2 --- Task completion and archival
      (End point: Apr 10):}\\
      \texttt{[2025-04-10, Group 2, \#8]} Luhao Zhao:
      ``\dots  the task of
      "Design System Overall Information Architecture
      and Site Map" has been \textcolor{blue}{successfully completed
      today!}\dots related content archived in Confluence\dots''

  \end{casebox}
  \vspace{-0.5em}
    \centering
    \begin{casebox}{Temporal Duration --- Working Days}
      \textbf{\mytriangle\ Question:}\\
      How many working days did it actually take Jiahui Zhao to plan
      the energy consumption baseline calculation logic for the
      enterprise energy consumption monitoring system and complete the
      design document based on the STL decomposition model?

      \vspace{0.1cm}
      \textbf{\mytriangle\ Correct Answer:}\\
      The task actually took 5 working days.

      \vspace{0.1cm}
      \textbf{\mytriangle\ Typical Wrong Answer:}\\
      ``7 working days'' (calendar-day count), ``6 working days''
      (off-by-one), or ``NOT ENOUGH INFORMATION.''

      \vspace{0.1cm}
      \textbf{\mytriangle\ Evidence}
      \texttt{(Full context: 2025-04-22 to 2025-04-28, Group 2)}:

      \textit{Anchor 1 --- Task initiation (Start point: Apr 22, Tue):}\\
      \texttt{[2025-04-22, Group 2, \#3]} Jiahui Zhao:
      ``\dots I will \textcolor{blue}{start} designing the
      \textcolor{blue}{energy consumption baseline calculation logic}
      today,\dots''

      \textit{Anchor 2 --- Task completion
      (End point: Apr 28, Mon):}\\
      \texttt{[2025-04-28, Group 2, \#12]} Jiahui Zhao:
      ``\dots the design task for the \textcolor{blue}{energy consumption
      baseline calculation logic has been completed}.
      The final version of the design document has been uploaded
      to Confluence\dots based on the \textcolor{blue}{STL decomposition
      model}\dots''

      \textit{Reasoning gap:} Apr~22 (Tue) to Apr~28 (Mon)
      spans 7 calendar days, but the question asks for
      \textit{working days}---the model must exclude the weekend
      (Apr~26 Sat, Apr~27 Sun) to arrive at the correct answer
      of \textbf{5}.

  \end{casebox}
  \vspace{-0.8em}
    \caption{Temporal Duration examples in Fine-grained Recall: calendar-day counting (top) and working-day counting (bottom).}
    \label{fig:recall_cases_part3}
\end{figure*}


\subsection{Memory Awareness Examples}\label{appendix:example-awareness}

As defined in \S3.1, Memory Awareness tests whether a system can understand stored information and apply it to novel scenarios. In this subsection, we illustrate each sub-task with a representative case and analyze how current systems fail to generalize memorized knowledge to unseen situations, distinguishing systems that have merely stored information from those that can act on it.

  \paragraph{Constraint.}                                                                                                                            
  This case presents a scenario---a schema evolution request (adding                          
  \texttt{workOrderNumber} to the alarm DTO)---that never appeared in the original conversation.
  The system must generalize implicit organizational norms from prior dialogue to determine
  responsibility, when the instruction is under-specified but not rule-conflicting. This evaluates
  \textit{Memory Awareness (Constraint)}: can the system extract and apply latent role relationships
  to reason about a novel situation?
  As shown in Figure~\ref{fig:mem_aware1}, the key distinction is between the \textit{data contract
  owner} (Ruiqing Jiang, who defined the DTO schema, 3--4 messages) and the \textit{data consumer}
  (Xuexin Yin, who builds notifications atop that schema, 30+ messages). Memory systems perform
  associative retrieval (``who is mentioned alongside alarm notifications?'') rather than causal
  attribution (``who must act first in the dependency chain?''), selecting the most \textit{visible}
  actor over the causally \textit{responsible} one. This is amplified by frequency bias: the query
  term ``alarm notification data'' overlaps semantically with Xuexin Yin's dense daily updates, while
  Ruiqing Jiang's sparse but decisive evidence uses different terminology (``alarm event DTO''),
  causing both vector-based and keyword-based retrieval to bury it.
\vspace{-1em}
\begin{figure*}[t!]
    \centering
    \begin{casebox}{Constraint}
      \textbf{\mytriangle\ Question:}\\
      \textcolor{blue}{\texttt{[User Request]}}\\
      Operations has a request to integrate the energy consumption
      monitoring and work order systems. High-level alarms need to
      automatically generate maintenance work orders. This
      definitely means adding a field like ``workOrderNumber'' to
      the alarm notification data. I'm not sure who's responsible
      for this$\dots$\\
      \textcolor{blue}{\texttt{[Options]}}
      \begin{enumerate}[label=\Alph*., topsep=0pt,
        leftmargin=1.4em, partopsep=0pt, itemsep=0pt]
          \item Should be evaluated and executed by Ruiqing
            Jiang$\dots$ she was responsible for defining the
            data structure of the alarm event object
            (DTO)$\dots$
          \item Should be submitted to Jian Wang, Head of Data
            Governance$\dots$ all changes to cross-service DTOs
            must undergo unified review$\dots$
          \item It should be led by Xuexin Yin$\dots$ the
            end-to-end owner for all subsequent iterations of
            the alert notification service's features$\dots$
          \item It should be handled by Xuexin Yin. He is the
            developer of the alarm notification service$\dots$
      \end{enumerate}

      \vspace{0.1cm}
      \textbf{\mytriangle\ Correct Answer:}\\
      A. Ruiqing Jiang ---  the
      architectural authority over the alarm event DTO: she
      defined the schema, delivered it to the downstream
      consumer.\\
      \vspace{0.1cm}
      \textbf{\mytriangle\ Typical Wrong Answer:}\\
      C or D (both selecting Xuexin Yin). Option~C fabricates
      a September~12 meeting that never occurred; Option~D
      contains true facts but conflates \textit{consuming}
      the data with \textit{defining} it---the downstream
      consumer cannot lead a schema change that must originate
      upstream.

      \vspace{0.1cm}
      \textbf{\mytriangle\ Evidence}
      \texttt{(Full context: 2025-09-05 to 2025-09-15, Group 2)}:

      \textit{Anchor 1 --- Dependency chain and schema delivery:}\\
      \texttt{[2025-09-05\,\&\,09-08, Group~2]} Xuexin Yin (twice):
      ``$\dots$please synchronize the DTO model$\dots$ as \textcolor{blue}{my notification template design depends on it}.''
      \texttt{[2025-09-08, Group~2, \#10]} Ruiqing Jiang:
      ``The \textcolor{blue}{alarm event DTO} on my side has been
      finalized$\dots$ I just sent it to your
      email \dots.''\\
      \texttt{[2025-09-08, Group~2, \#11]} Xuexin Yin:
      ``Received$\dots$ I can start writing the core logic
      for the notification service \textcolor{blue}{based on this
      model} \dots.''\\
  \textit{Anchor 2 --- Precedent:} Xuexin asked about adding \texttt{info} to the alert-level enum; Ruiqing made the unilateral decision
  (\texttt{[09-09]})---structurally similar to the benchmark question.
  \end{casebox}
  \vspace{-0.8 em}
    \caption{Constraint example in Memory Awareness.}
    \label{fig:mem_aware1}
\end{figure*}

\paragraph{Proactivity.}                                                                                                                           
  This case introduces a technically compelling but \textbf{non-compliant} request---adopting Zustand
  for a new dashboard module, bypassing the project's standardised technology stack. The system must                                                 
  proactively recall an explicit organisational rule and flag the conflict, even when the request is                                                 
  framed as urgent and pragmatically reasonable. This evaluates \textit{Memory Awareness
  (Proactivity)}: can the system detect and resist violations of established policies under persuasive
  framing?
  As shown in Figure~\ref{fig:mem_aware2}, the correct answer requires recalling that Redux Toolkit
  was formally selected as the unified standard---approved by the project lead, integrated with CI
  verification, and documented on Confluence. When retrieval fails to surface this decision, 76\% of
  systems \emph{fabricate organisational consent}, attributing to the project lead a ``per-module
  freedom'' policy that never existed. This \textit{policy hallucination} is amplified by
  technical-merit bias: Zustand \emph{is} lighter and faster to develop with, making the proposal
  sound reasonable on engineering merits alone. Without the factual anchor of the recorded team
  decision, systems default to evaluating technical appeal rather than checking against institutional
  policy.

\begin{figure*}[t!]
    \centering
    \begin{casebox}{Proactivity}
      \textbf{\mytriangle\ Question:}\\
      \textcolor{blue}{\texttt{[User Request]}}\\
      The new ``Real-time Alert Dashboard'' for the energy
      consumption system is a high-priority request; they want
      to see something by next week. I've evaluated it, and our
      current Redux Toolkit setup is too heavy and slow for
      development. I plan to go directly with Zustand; it's
      lightweight and quick to get started with. Please make a
      note that Zustand will be the technology choice for this
      module.\\
      \textcolor{blue}{\texttt{[Options]}}
      \begin{enumerate}[label=\Alph*., topsep=0pt,
        leftmargin=1.4em, partopsep=0pt, itemsep=0pt]
          \item Not recommended. According to technical
            guidelines from Frontend Architect Jianbo Wang$\dots$
            prioritise React's native Context API$\dots$
          \item Yes. Project lead Mingzhi Li mentioned$\dots$
            the development team can choose a suitable state
            management library based on actual circumstances$\dots$
          \item Yes, a perfect opportunity for a pilot.
            At the technical selection review meeting on
            June~15$\dots$ Zustand could be piloted in this
            module$\dots$
          \item Not recommended. Based on the technical
            selection at inception, the system has uniformly
            integrated Redux Toolkit$\dots$ all new features
            should adhere to this standard. Usage examples
            are on Confluence$\dots$
      \end{enumerate}

    \textbf{\mytriangle\ Correct Answer:}\\                                                                                                                      
        D. Redux Toolkit --- the dialogue records a complete decision-implementation-documentation                                                            
        lifecycle. Redux Toolkit was approved by the project lead, CI-integrated, and documented                                                           
        on Confluence, constituting a formalised project standard.                                                                                                   

        \vspace{0.1cm}
        \textbf{\mytriangle\ Typical Wrong Answer:}\\
        B (76.0\%) --- fabricates a ``per-module flexibility'' policy, attributing to the project
        lead an authorisation that never existed. Option~C (5.3\%) further fabricates a specific
        ``June~15 review meeting'' approving a Zustand pilot.

      \vspace{0.1cm}
      \textbf{\mytriangle\ Evidence}
      \texttt{(Full context: 2025-05-30 to 2025-06-05, Group 2)}:

      \textit{Anchor 1 --- Technology selection decision:}\\
      \texttt{[2025-05-30, Group~2, \#3]} Yanjun Fan:
      ``For the state management library, \textcolor{blue}{I'll
      choose Redux Toolkit} and integrate and configure it
      next Monday.''\\
      \texttt{[2025-05-30, Group~2, \#6]} Mingzhi Li
      (project lead) approves the plan without objection.

      \textit{Anchor 2 --- Formalisation and documentation:}\\
      \texttt{[2025-06-05, Group~2, \#5]} Yanjun Fan:
      ``The task of `Building React Project Structure and
   Integrating State Management Library' has
      been completed. The code has been merged$\dots$ CI
      pipeline passed. \dots The documentation on
      Confluence has been updated, including$\dots$
      \textcolor{blue}{examples of state management usage}.''
  \end{casebox}
    \caption{Proactivity example in Memory Awareness.}
    \label{fig:mem_aware2}
\end{figure*}

\paragraph{Update.}
  This case requires the system to recognise that a \textbf{base rule has been overridden} by a
  later directive and to \textbf{chain the override with its operational definition}. The team
  originally adopted GitFlow, establishing the standard emergency fix procedure: create a
  \texttt{hotfix} branch from \texttt{master} (Option~A). Forty-seven days later, a new directive
  maps \texttt{CORE\_ALGO} P0 defects to a dedicated emergency plan
  (\texttt{PROC\_OVERRIDE\_RED\_V1}); then, a separate message defines the plan's
  operational steps (rollback~+ \texttt{critical\_fix} branch). The system must model this rule
  evolution over time and chain three separately stored pieces---base rule, override mapping, and
  override definition---into a coherent action sequence. This evaluates \textit{Memory Awareness
  (Update)}: can the system resolve temporal precedence and correctly compose amendments with the
  base rule?
  As shown in Figure~\ref{fig:mem_aware3}, all memory systems successfully retrieve the override
  mapping (Anchor~2)---the plan name (\texttt{PROC\_OVERRIDE\_RED\_V1})---but none retrieve
  Anchor~3: its operational definition (rollback~+ \texttt{critical\_fix} branch), stored  in a different thread. The dominant distractor (Option~B, 90.3\%) exploits this partial
  retrieval by pairing the \emph{correct plan name} with a \emph{fabricated definition} (``online
  isolation and repair''), creating a half-right trap that confirms the model's incomplete evidence
  while supplying the missing piece. Even the oracle configuration falls for this trap, indicating
  the failure extends beyond retrieval to a reasoning vulnerability against plausible-sounding
  fabrications.

\begin{figure*}[t!]
    \centering
    \begin{casebox}{Update}
      \textbf{\mytriangle\ Question:}\\
    \textcolor{blue}{\textcolor{blue}{\texttt{[User Request]}}}\\
      Senior Engineer Wang is troubleshooting an online
      P0-level defect originating from the core investment
      advisory algorithm. The JIRA ticket's \texttt{CATEGORY}
      is marked \texttt{CORE\_ALGO}. According to the team's
      R\&D guidelines, how should he formulate the fix
      strategy?\\
      \textcolor{blue}{\textcolor{blue}{\texttt{[Options]}}}
      \begin{enumerate}[label=\Alph*., topsep=0pt,
        leftmargin=1.4em, partopsep=0pt, itemsep=0pt]
          \item Follow \texttt{STANDARD\_PROC\_ENFORCED}$\dots$
            create a standard \textcolor{blue}{\texttt{hotfix}} branch from
            \texttt{master}$\dots$ ensure CI/CD compatibility
            and audit-log integrity$\dots$
          \item This triggered
            \texttt{PROC\_OVERRIDE\_RED\_V1}$\dots$ address
            core risks through ``online isolation and
            repair''$\dots$ deploy fix to isolated nodes for
            canary testing$\dots$
          \item Trigger
            \texttt{PROC\_OVERRIDE\_BLUE\_V1} degradation
            plan$\dots$ service degradation on the algorithm
            module, return safe fallback data$\dots$
          \item Immediately suspend all standard procedures.
            Trigger rollback to the most recent stable Tag,
            then create a \textcolor{blue}{\texttt{critical\_fix}} branch for
            full-cycle remediation$\dots$
      \end{enumerate}
     \textbf{\mytriangle\ Correct Answer:}\\
      \vspace{0.1cm}
        D. Rollback~+ \texttt{critical\_fix} --- requires chaining two directives:
        \texttt{CORE\_ALGO} P0 $\rightarrow$ \texttt{PROC\_OVERRIDE\_RED\_V1} $\rightarrow$
        rollback to stable Tag and start a \texttt{critical\_fix} branch.\\
        \vspace{0.1cm}
        \textbf{\mytriangle\ Typical Wrong Answer:}\\
        B (90.3\%) --- names the correct plan (\texttt{PROC\_OVERRIDE\_RED\_V1}) but
        fabricates its definition as ``online isolation and repair,'' creating a half-right
        trap that exploits partial retrieval of Hop~1 without Hop~2.\\
      \vspace{0.1cm}
      \textbf{\mytriangle\ Evidence}
      \texttt{(Full Context: 2025-05-14 to 2025-07-10, Group 3)}:\\
      \textit{Anchor 1 --- Base rule (the standard emergency process):}\\
      \texttt{[2025-05-14, Group~3, \#2]} Haitao Cao:
      ``$\dots$ use \textcolor{blue}{GitFlow} as our branching model,
      creating master, develop, feature, release, and
      \textcolor{blue}{hotfix} branches$\dots$''\\
      \textit{Anchor 2 --- Override mapping (Hop~1, 47~days later):}\\
      \texttt{[2025-06-30, Group~3, \#6]} Mingzhi Li:
      ``$\dots$all P0-level defects in the JIRA system with
      \texttt{CATEGORY} as \textcolor{blue}{\texttt{CORE\_ALGO}}
      must be mandatorily linked to and trigger the
     \textcolor{blue}{\texttt{PROC\_OVERRIDE\_RED\_V1}} emergency
      plan. This regulation is effective immediately.''\\
      \textit{Anchor 3 --- Override definition (Hop~2):}\\
      \texttt{[2025-07-10, Group~3, \#7]} Haitao Cao:
      ``After \texttt{PROC\_OVERRIDE\_RED\_V1} is triggered:
      1.~\textcolor{blue}{Immediately terminate} the current
      standardised emergency recovery process;
      2.~Trigger a \textcolor{blue}{rollback}$\dots$  Tag;
      3.~Automatically start a
      \textcolor{blue}{\texttt{critical\_fix} branch} for
      subsequent full-cycle repairs.''
    
  \end{casebox}
    \caption{Update example in Memory Awareness.}
    \label{fig:mem_aware3}
\end{figure*}


\subsection{Profile Understanding Examples}\label{appendix:example-profile}
As defined in \S3.1, Profile Understanding evaluates whether a memory system can infer stable, implicit persona traits from long-term dialogue. In real-world settings, users rarely articulate how they communicate or what professional skills they possess; these traits are expressed implicitly and consistently across many conversations. A capable memory system should retain a user's dialogue history and distil from it the recurring patterns---communication style, skill boundaries, professional role---that constitute the user's persona. In this subsection, we present one case per sub-task and analyze how current systems fail to capture such implicit persona signals. For brevity, each case's evidence focuses on the factual anchors needed to answer the question. Because the persona dimensions are implicitly reflected throughout each user's long-term dialogue history, rather than reproducing those dialogue excerpts, we
  directly provide the ground-truth profile (showing only the dimension under evaluation) in each figure.  Readers who wish to verify the persona--dialogue alignment can consult the complete dialogue files in the released dataset.

\vspace{-1em}
\paragraph{Style.}
This case tests whether the memory system can capture a speaker's \textbf{communication style}
beyond factual content. The user (Ruiqing Jiang) asks the system to draft a reply on her behalf
regarding the alarm-level DTO design. Two of the four options (B and~C) are \textit{factually
identical}---both correctly state that only \texttt{warning} and \texttt{critical} are defined,
with \texttt{info} reserved---but differ \textit{solely in tone}. The system must leverage
historical dialogue to align with Ruiqing Jiang's characteristic informal, emoji-rich register.
This evaluates \textit{Profile Understanding (Style)}: can the system reconstruct not just \textit{what}
someone knows, but \textit{how} she communicates?
As shown in Figure~\ref{fig:profile_un2}, this case isolates \textit{style} as the sole discriminating
dimension between the correct and dominant wrong answer.
(1)~\emph{Style retrieval blind spot}: All memory systems successfully retrieved the factual content
of Ruiqing Jiang's messages about the alarm DTO. Yet these same source messages also carry her
characteristic style markers---\emoji{rocket} appended to daily updates, ``Good question!'' as a
catchphrase, \emoji{wink} in informal exchanges.%
The retrieval and summarization pipeline treats
such  features as non-semantic noise, stripping them while preserving only the
informational payload. The style evidence is \textit{co-located with the retrieved facts} but
invisible to the downstream LLM.
(2)~\emph{Default persona bias}: Receiving only dry technical fragments, the LLM falls back to a
``standard AI assistant'' register---formal, passive-voice, personality-free. Option~C is the
prototypical output of this default mode. That every memory-augmented configuration across all
tested LLMs converges on the same wrong answer confirms the failure is \textit{systemic} (rooted
in the retrieval pipeline) rather than model-specific.
(3)~\emph{Fact--style decoupling as a diagnostic}: Because Options~B and~C are factually identical
and differ only in register, this question functions as a controlled experiment: selecting~C over~B
demonstrates successful \textit{factual recall} coupled with failed \textit{persona replication}.
The ground-truth profile (\texttt{Emoji\_Usage:~Frequent}, \texttt{Formality:~Casual}) independently
confirms that the casual, emoji-rich tone of~B reflects a stable personality trait rather than an
anomalous outlier.

\begin{figure*}[t!]
    \centering
    \begin{casebox}{Style}
      \textbf{\mytriangle\ Question:}\\
      \textcolor{blue}{\texttt{[User Request]}}\\
      I am Ruiqing Jiang, an algorithm engineer for the energy
      consumption monitoring system. My colleague Xuexin Yin asked
      me in the group chat about the design of the alarm level
      DTO. Please help me draft a reply.\\
      \textcolor{blue}{\texttt{[Options]}}
      \begin{enumerate}[label=\Alph*., topsep=0pt,
        leftmargin=1.4em, partopsep=0pt, itemsep=0pt]
          \item @Xuexin Yin Hello, regarding the design of alert
            levels, the current DTO already defines \textcolor{blue}{three levels:
            `warning', `critical', and `info'}. Please ensure full
            support$\dots$
          \item @Xuexin Yin Good question! The DTO currently only
            has \textcolor{blue}{`warning' and `critical'}. I'll add `info' as a
            reserved item to the documentation. For now, you can
            develop with the existing two and add a default case
            as a fallback \emoji{wink}
          \item @Xuexin Yin Hello, regarding the design of the
            alarm level DTO, currently \textcolor{blue}{only `warning' and
            `critical' levels} are defined. Considering future
            scalability, the `info' level will be noted as a
            reserved item. It is recommended that you develop
            based on the existing levels first and set up default
            handling logic.
          \item @Xuexin Yin Good question! I specifically designed
            this DTO with extensibility in mind, so from the
            beginning, I included \textcolor{blue}{three levels$\dots$} \emoji{flexed-biceps}
      \end{enumerate}
      \vspace{0.1cm}
   \textbf{\mytriangle\ Correct Answer:}\\
        B --- the only option satisfying both \textit{factual} and \textit{stylistic} constraints:
        correct fact (two levels defined, \texttt{info} reserved) delivered in Ruiqing Jiang's
        characteristic casual, emoji-rich register (``Good question!'' opener, \emoji{wink} closer).

  \textbf{\mytriangle\ Typical Wrong Answer:}\\                                                                         C (selected by \textit{all} memory system) --- factually correct but drafted in a generic tone rather than matching   Ruiqing Jiang's personalised communication style. Options~A and~D additionally fabricate incorrect facts.\\       
      \vspace{0.1cm}
      \textbf{\mytriangle\ Evidence}
      \texttt{(Full context: 2025-09-05 to 2025-09-09, Group 2)}:\\
      \textit{Anchor 1 --- Factual ground truth (alert-level enum):}\\
      \texttt{[2025-09-09, Group~2]} Ruiqing Jiang:
      ``@Xuexin Yin Good question! The current design only
      includes \textcolor{blue}{`warning' and `critical'} for now.
     \dots I will
      \textcolor{blue}{note and reserve} the possibility of
      `info'$\dots$ You can proceed with these two levels
      for now and just add a default logic. \emoji{wink}''\\
      \textbf{\mytriangle\ Profile}
      \texttt{(Ruiqing Jiang --- Communication Style)}:\\
          Formality: \textcolor{blue}{Casual};
          Verbosity: Concise;
          Humor: \textcolor{blue}{Frequent};
          Emoji Usage: \textcolor{blue}{Frequent};
          Directness: Direct;
          Warmth: Friendly;
          Questioning Style: Probing
  \end{casebox}
    \caption{Style example in Profile Understanding.}
    \label{fig:profile_un2}
\end{figure*}

  \begin{figure*}[t!]                                                                                          
      \centering                                                                                                                                                     
      \captionsetup{skip=2pt}                                                                                                                                        
      \begin{casebox}{Skill}
        \textbf{\mytriangle\ Question:}\\
        \textcolor{blue}{\texttt{[User Request]}}\\
        I (Jie Gu) have just joined the ``Data Asset Catalog''
        project team. Boss Zhang has asked me to focus on the
        technical solutions for data lineage and metadata
        collection. I've reviewed the project goals, and the core
        objective is to solve the problem of business users finding
        it difficult to locate and understand data. \textcolor{blue}{Mingzhi Li
        previously proposed} several open-source solutions. Now \textcolor{blue}{I
        need to provide an initial technical recommendation.} What
        would you suggest?\\
        \textcolor{blue}{\texttt{[Options]}}
        \begin{enumerate}[label=\Alph*., topsep=0pt,
          leftmargin=1.4em, partopsep=0pt, itemsep=0pt]
            \item $\dots$\textcolor{blue}{Amundsen} has clear advantages in
              data discovery$\dots$ Although it uses a
              \textcolor{blue}{Python} tech stack, it's now deployed as
              microservices, so we can independently deploy its
              service clusters and integrate with our backend
              via REST API$\dots$
            \item $\dots$we should abandon the batch processing
              solution and instead develop an in-house real-time
              metadata capture system deeply integrated with our
              existing {Flink} streaming
              architecture$\dots$ build and update the lineage
              graph in memory using the {Gelly} graph
              computing library$\dots$
            \item $\dots$I recommend focusing on evaluating
              \textcolor{blue}{Marquez}. It is inherently part of the
              \textcolor{blue}{Java} ecosystem, allowing for seamless
              integration with our existing \textcolor{blue}{Spring Boot}
              technology stack$\dots$ leverage its support for
              the OpenLineage standard to standardize metadata
              collection$\dots$
            \item $\dots$consider using the {Python}
              ecosystem to quickly prototype$\dots$ use
              {Faust}, a Python stream processing
              library$\dots$ develop the API service using
              {FastAPI}$\dots$
        \end{enumerate}
        \textbf{\mytriangle\ Correct Answer:}\\
        {C}.Marquez --- the only Java-native, Spring Boot-based option,
        simultaneously matching Jie Gu's personal skill boundary
        (``Java expert'') and the team's ratified backend stack.\\
        \textbf{\mytriangle\ Typical Wrong Answer:}\\
       { A. }(majority of failing configurations) --- selects Amundsen
        for its industry reputation, reflecting generic best-practice
        reasoning that ignores both the recommender's Java expertise
        and the team's technology decision. Options~B and~D fabricate
        in-house development mandates, contradicting the open-source
        evaluation scope.\\
        \textbf{\mytriangle\ Evidence}
        \texttt{(Full context: 2025-01-16 to 2025-02-10, Groups 1--2)}:\\
        \textit{Anchor 1 --- Team tech-stack ratification:}\\
        \texttt{[2025-02-10, Group~2]} Mingzhi Li:
        ``$\dots$The report's conclusion leans towards choosing
        \textcolor{blue}{Java/Spring Boot as the backend technology
        stack}$\dots$ ''\\
        \textit{Anchor 2 --- Open-source candidates:}\\
        \texttt{[2025-01-16, Group~2]} Mingzhi Li:
        ``$\dots$We've  researched a few open-source
        data lineage  tools, such as
        \textcolor{blue}{Marquez and Amundsen}$\dots$''\\
        \textbf{\mytriangle\ Profile}
        \texttt{(Jie Gu --- Skills)}:
            \textcolor{blue}{Java~(\underline{strong})};
            Spring Boot~(\underline{medium});
            MySQL~(medium);
            Kubernetes~(low);
            \textit{no Python skill listed}
    \end{casebox}
     \vspace{-4pt}
      \caption{Skill example in Profile Understanding.}
      \label{fig:profile_un3}
  \vspace{-4pt}
      \centering
      \captionsetup{skip=2pt}
      \begin{casebox}{Role}
        \textbf{\mytriangle\ Question:}\\
        \textcolor{blue}{\texttt{[User Request]}}\\
        \textcolor{blue}{I (Yutong Song)} heard that the team recently had a
        successful practice in \textcolor{blue}{ensuring data accuracy}, which
        resolved the issue of inaccurate data on the large screen
        dashboards. My manager wants me to \textcolor{blue}{summarize this case and
        propose some subsequent systematic improvement suggestions}.
        Please help me prepare a response.\\
        \textcolor{blue}{\texttt{[Options]}}
        \begin{enumerate}[label=\Alph*., topsep=0pt,
          leftmargin=1.4em, partopsep=0pt, itemsep=0pt]
            \item $\dots$the root cause is a {performance
              bottleneck} in the real-time calculation of
              ``Integrated Energy Efficiency Ratio''$\dots$
              design a set of dedicated {performance test
              cases}$\dots$ output a detailed performance test
              report$\dots$
            \item $\dots$the root cause is a {performance bottleneck}
              $\dots$ optimize the SQL$\dots$ introducing a
              caching layer like {Redis}$\dots$ add
              monitoring and alerts for the API query's response
              time$\dots$
            \item $\dots$identified and solved through
              \textcolor{blue}{cross-validation}$\dots$ 1.~Sort out the
              complete \textcolor{blue}{data pipeline}$\dots$ 2.~Add
              automated data verification scripts$\dots$
              3.~\textcolor{blue}{Optimize relevant API designs} from a
              system architecture perspective$\dots$
            \item $\dots$Xinmeng Tian and Jiahui Zhao efficiently
              solved the problem through
              \textcolor{blue}{cross-validation}$\dots$ 1.~Formalize
              $\dots$into a set of \textcolor{blue}{standard test
              cases}$\dots$ 2.~Add$\dots$to the
              \textcolor{blue}{regression test suite}$\dots$
              3.~$\dots$updating the \textcolor{blue}{test process
              documentation}$\dots$
        \end{enumerate}
        \textbf{\mytriangle\ Correct Answer:}\\
        C. Correctly identifies the cross-validation incident and
        frames recommendations from a \textit{developer's} perspective
        (data-pipeline review, API optimization)---actions within
        Yutong Song's competence as a frontend/full-stack engineer.\\
        \textbf{\mytriangle\ Typical Wrong Answer:}\\
        D (nearly all configurations). Same correct incident, but framed
        from a \textit{QA} perspective (test cases, regression suites, test
        documentation)---responsibilities of the QA engineer who executed
        the cross-validation, not of Yutong Song. \\
        \textbf{\mytriangle\ Evidence}
        \texttt{(Full context:2025-04 to 2025-10, Groups 1--2)}:\\
        \textit{Anchor --- The cross-validation incident:}\\
        \texttt{[2025-10-24, Group~2]} Xinmeng Tian:
        ``@Jiahui Zhao \textcolor{blue}{Cross-validation complete},
        data fully matched! The accuracy of all charts and
        KPI data on the large screen has been verified.''\\
        \textbf{\mytriangle\ Profile}
        \texttt{(Yutong Song --- Title)}:
            \textcolor{blue}{Frontend / Full-stack Developer}
    \end{casebox}
      \vspace{-4pt}
      \caption{Role example in Profile Understanding.}
      \label{fig:profile_un4}
  \end{figure*}

\paragraph{Skill.}
This case tests whether the memory system can constrain a technical recommendation by the
\textbf{speaker's skill profile}. Jie Gu, a newly joined member of the ``Data Asset Catalog''
project, is asked to propose an open-source solution for data lineage. Multiple options are
technically viable, but only one aligns with both his personal expertise (\textit{Java specialist})
and the team's confirmed technology stack (\textit{Java/Spring Boot}). This evaluates
\textit{Profile Understanding (Skill)}: can the system select the option that a specific person
\textit{would} recommend, rather than the option that is generically ``best''?
As shown in Figure~\ref{fig:profile_un3}, this case demonstrates that technical recommendations must be
grounded in the recommender's identity.
(1)~\emph{Ignoring persona-skill constraints}: As shown in the ground-truth profile,
Jie Gu's skill set centers on Java (\texttt{Java:~strong}, \texttt{Spring Boot:~medium}) with
no Python skill listed---a trait implicitly but consistently reflected across his long-term work
dialogues (as noted at the beginning of this subsection, we provide the profile directly for
brevity rather than reproducing those dialogue excerpts). Memory systems that rely on topical
similarity (``data lineage,'' ``metadata collection'') retrieve feature comparisons but fail to
distil this persona-level constraint from the user's dialogue history, missing that \textit{whose
perspective} the answer should reflect is as important as the technical content being compared.
Without this constraint, the LLM defaults to a domain-expert mode and selects the most
feature-rich or industry-popular option (Amundsen).
(2)~\emph{Detachment from team tech-stack context}: Even when the persona signal is missed, the
team's formal adoption of Java/Spring Boot provides a second, independent constraint: any
recommendation introducing a Python-based system contradicts a ratified architectural decision.
Systems that select~A acknowledge the Python mismatch in the option text itself (``Although it uses
a Python tech stack$\dots$'') yet still choose it, indicating that retrieved context failed to supply
the countervailing organizational constraint.
\vspace{-1em}
\paragraph{Role.}
This case tests whether the memory system can align improvement suggestions with the
\textbf{speaker's professional role}. Yutong Song, a frontend/full-stack developer, is asked to
summarize a data-accuracy incident and propose systematic improvements. Two options (C and~D)
correctly describe the same incident (cross-validation), but frame the recommendations from
opposing professional perspectives: Option~C adopts a \textit{developer/architect} lens
(data-pipeline review, API optimization), while Option~D adopts a \textit{QA} lens (test cases,
regression suites, test documentation). This evaluates \textit{Profile Understanding (Role)}: can the
system infer that a developer's ``systematic improvement suggestions'' should concern code and
architecture, not testing processes?
As shown in Figure~\ref{fig:profile_un4}, this case layers event identification and role-appropriate
framing.
(1)~\emph{Successful fact retrieval, failed role inference}: The majority of systems correctly
retrieved the cross-validation incident (rejecting the Redis distractor), yet still chose~D over~C.
This reveals that the bottleneck is not in \textit{what happened} but in \textit{whose voice} the
summary should adopt. The memory pipeline retrieves event facts but does not propagate the speaker's
job function into the reasoning context.
(2)~\emph{Semantic attraction of ``cross-validation'' toward QA framing}: The term
``cross-validation'' carries a strong connotation of testing and verification. Combined with the
fact that the incident was \textit{executed} by a QA engineer (Xinmeng Tian), the retrieved evidence
naturally co-occurs with testing vocabulary, priming the LLM toward a QA-oriented response and
making Option~D appear as the ``natural continuation'' of the retrieved context.
(3)~\emph{Absence of negative reasoning}: Selecting~C over~D requires the system to perform
\textit{exclusionary} inference: a frontend/full-stack developer would \textit{not} propose
maintaining regression test suites or updating test process documentation. The ground-truth profile of the target person 
(\texttt{JavaScript:~strong}, \, \texttt{React:~medium}, \, \texttt{Node.js:~medium}; no QA skills)
confirms this boundary. Current memory architectures lack a mechanism to retrieve and apply such
\textit{negative} persona constraints.

\end{sloppy}
\end{document}